\documentclass[final]{cvpr}

\usepackage{times}
\usepackage{epsfig}
\usepackage{graphicx}
\usepackage{amsmath}
\usepackage{amssymb}
\usepackage{booktabs}
\usepackage{caption}
\usepackage{subcaption}
\usepackage{multirow}

\usepackage[pagebackref=true,breaklinks=true,colorlinks,bookmarks=false]{hyperref}

\def\confYear{CVPR 2021}

\begin{document}

\title{Learning Compositional Representation for 4D Captures with Neural ODE}

\author{Boyan Jiang$^{1*}$ \quad Yinda Zhang$^{2*}$ \quad Xingkui Wei$^{1}$ \quad Xiangyang Xue$^{1}$ \quad Yanwei Fu$^{1}$ \\
$^{1}$ Fudan University \quad $^{2}$ Google
}

\maketitle

\pagestyle{empty}
\thispagestyle{empty}

{\let\thefootnote\relax\footnotetext{$^{*}$ indicates equal contributions.}}
{\let\thefootnote\relax\footnotetext{Boyan Jiang  Xingkui Wei and Xiangyang Xue are with the  School of Computer Science, Fudan University.}}
{\let\thefootnote\relax\footnotetext{Yanwei Fu is with the 
School of Data Science, MOE Frontiers Center for
Brain Science, and Shanghai Key Lab of Intelligent Information Processing, Fudan University.}}

\begin{abstract}
Learning based representation has become the key to the success of many computer vision systems. While many 3D representations have been proposed, it is still an unaddressed problem how to represent a dynamically changing 3D object. In this paper, we introduce a compositional representation for 4D captures, i.e. a deforming 3D object over a temporal span, that disentangles shape, initial state, and motion respectively.
Each component is represented by a latent code via a trained encoder.
To model the motion, a neural Ordinary Differential Equation (ODE) is trained to update the initial state conditioned on the learned motion code, and a decoder takes the shape code and the updated state code to reconstruct the 3D model at each time stamp.
To this end, we propose an Identity Exchange Training (IET) strategy to encourage the network to learn effectively decoupling each component.
Extensive experiments demonstrate that the proposed method outperforms existing state-of-the-art deep learning based methods on 4D reconstruction, and significantly improves on various tasks, including motion transfer and completion.

\end{abstract}

\section{Introduction}

\begin{figure}[t]
\centering \includegraphics[width=0.95\linewidth]{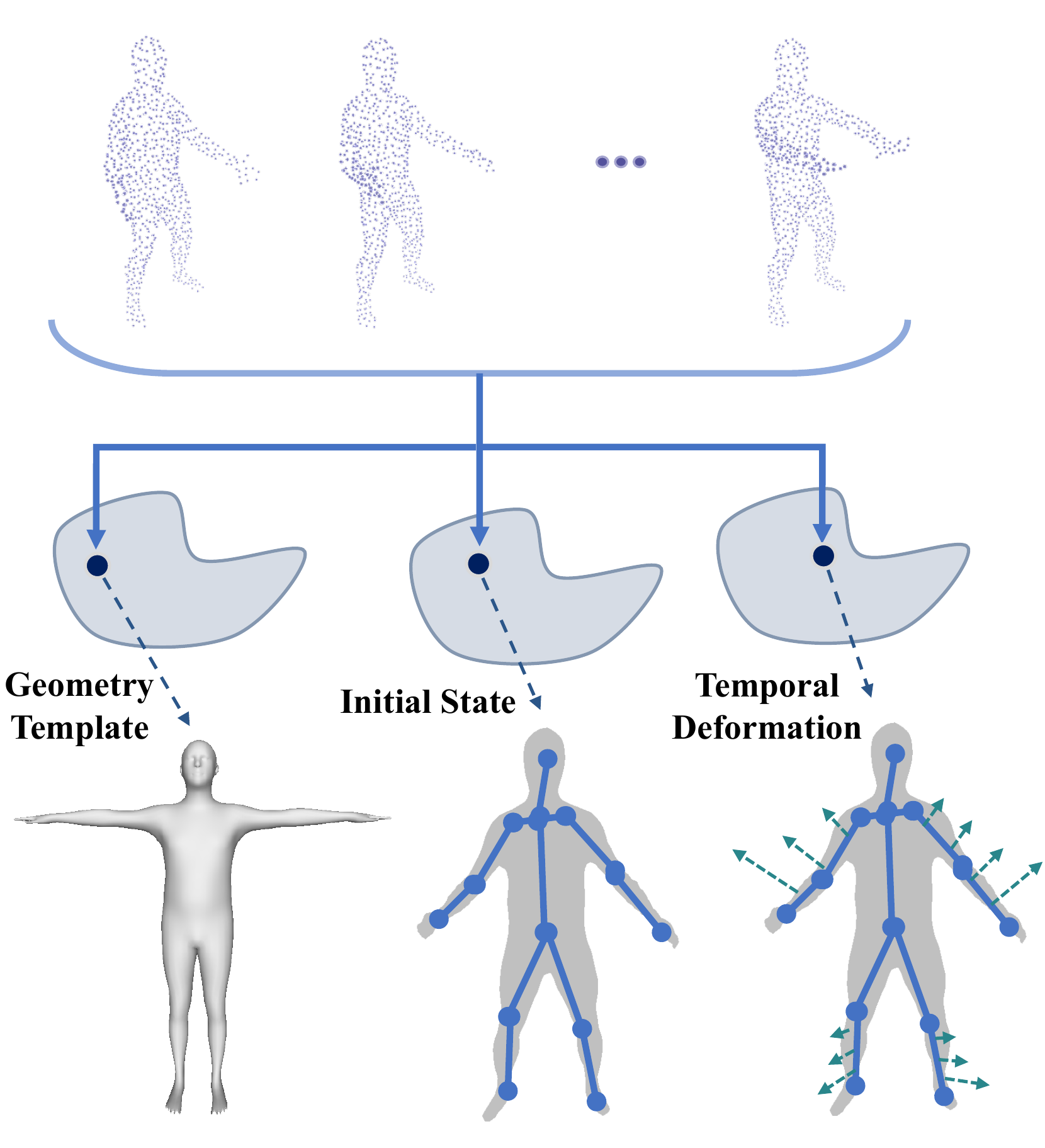}
\caption{We present a compositional representation for 4D object dynamics, through which the input point cloud sequence is disentangled into semantically meaningful representations in three latent spaces for geometry template, initial state, and temporal deformation.}
\vspace{-0.15in}
\label{fig:titlepage} 
\vspace{-0.1in}
\end{figure}

Shape representation is one of the core topics in 3D computer vision, especially in the era of deep learning.
Early work uses explicit representation, e.g. volume \cite{choy20163d,GirdharFRG16,wang2017cnn}, point cloud \cite{qi2016pointnet,fan2017point,QiLWSG18,achlioptas2017representation}, and mesh \cite{groueix2018atlasnet,kanazawa2018learning,pixel2mesh} for 3D related tasks, such as shape reconstruction, synthesis, and completion.
Recently, deep implicit representation \cite{Occupancy_Networks, park2019deepsdf,jiang2020local} shows promising performance in producing accurate geometry with appealing surface details.
However, arguably, we, humans, stay in a 3D world with an additional temporal dimension, and the majority of data we perceive everyday are moving or deforming 3D objects and scenes.
Many existing applications also require understanding or reconstruction of 4D data, such as autonomous driving, robotics, and virtual or augmented reality.
But the deep representation for 4D data, i.e. a deforming 3D object over a time span, is barely missing in the literature.
As a pioneer work, Niemeyer \etal \cite{niemeyer2019occupancy} propose to predict velocity field of the 3D motion via a Neural ODE \cite{chen2018neural}.
However, the method mainly focuses on recovering and integrating local flow for 4D reconstruction, which might accumulate error and thus produce sub-optimal quality.

In this work, we propose a novel deep compositional representation for 4D captures. This representation can be used to reconstruct 4D captures, and it also extracts key understanding that supports high-level tasks, such as motion transfer, 4D completion, or future prediction.
This is achieved by an encoder that takes a 4D capture as input and produces latent codes representing the geometry template, initial state, and temporal deformation respectively. 
Taking human as an example, these three key factors are commonly understood as the identity, initial body pose, and motion\footnote{Since the experiment is mainly conducted on 4D human captures, we use these terms interchangeably.}.

To reconstruct the 4D capture, we design a novel architecture taking three latent codes as inputs.
First, we keep the geometry template code (i.e. the identity) unchanged over time since it is not affected by the motion.
Then, we propose a novel conditional latent Neural ODE to update the initial state code (i.e. the initial body pose) conditioned on the deformation code (i.e. the motion).
The temporally varying state code is further concatenated with the geometry template code, and fed into a decoder to reconstruct an implicit occupancy field for each time frame, which recovers the 3D shape over time.
Mostly similar to us, Occupancy Flow \cite{niemeyer2019occupancy} also use Neural ODE \cite{chen2018neural} to update the position of each 3D point for 4D reconstruction.
In contrast, our method applies the Neural ODE to update the latent state code that controls the shape globally, which is empirically more stable.

To learn our compositional representation, we propose a training strategy to enable the encoder to decouple the geometry template and deformation, inspired by He \etal \cite{he2018probabilistic}.
Specifically, we take two 4D captures from different subjects and extract their latent codes respectively. We then swap their geometry template code and train the network to reconstruct the motion with swapped geometry template.
The training is fully supervised by synthetic data, where the parametric model is used to generate 4D captures with the same motion but different geometry template, e.g. SMPL model \cite{loper2015smpl} for humans.
We found this training strategy is effective in separating geometry template from the motion, which naturally supports motion transfer.
The representation also enables 4D completion from captures with either missing frames or partial geometry by solving an optimization to update the latent codes until the partial observation is best explained.

Our contributions can be summarized as follows. First, we design a novel deep representation for 4D captures that understands the geometry template, initial state, and temporal deformation, and propose a novel training strategy to learn it.
Second, we propose a novel decoder to reconstruct 4D captures from the learned representation, which includes, as a key component, a conditional Neural ODE to recover varying pose codes under the guidance of the motion code; and these codes are then translated into an occupancy field in implicit representation to recover the varying shape.
Finally, we show that our model outperforms state-of-the-art methods on 4D reconstruction, and our compositional representation is naturally suitable for various applications, including motion transfer and 4D completion. 


\begin{figure*}[t]
\centering \includegraphics[width=0.95\linewidth]{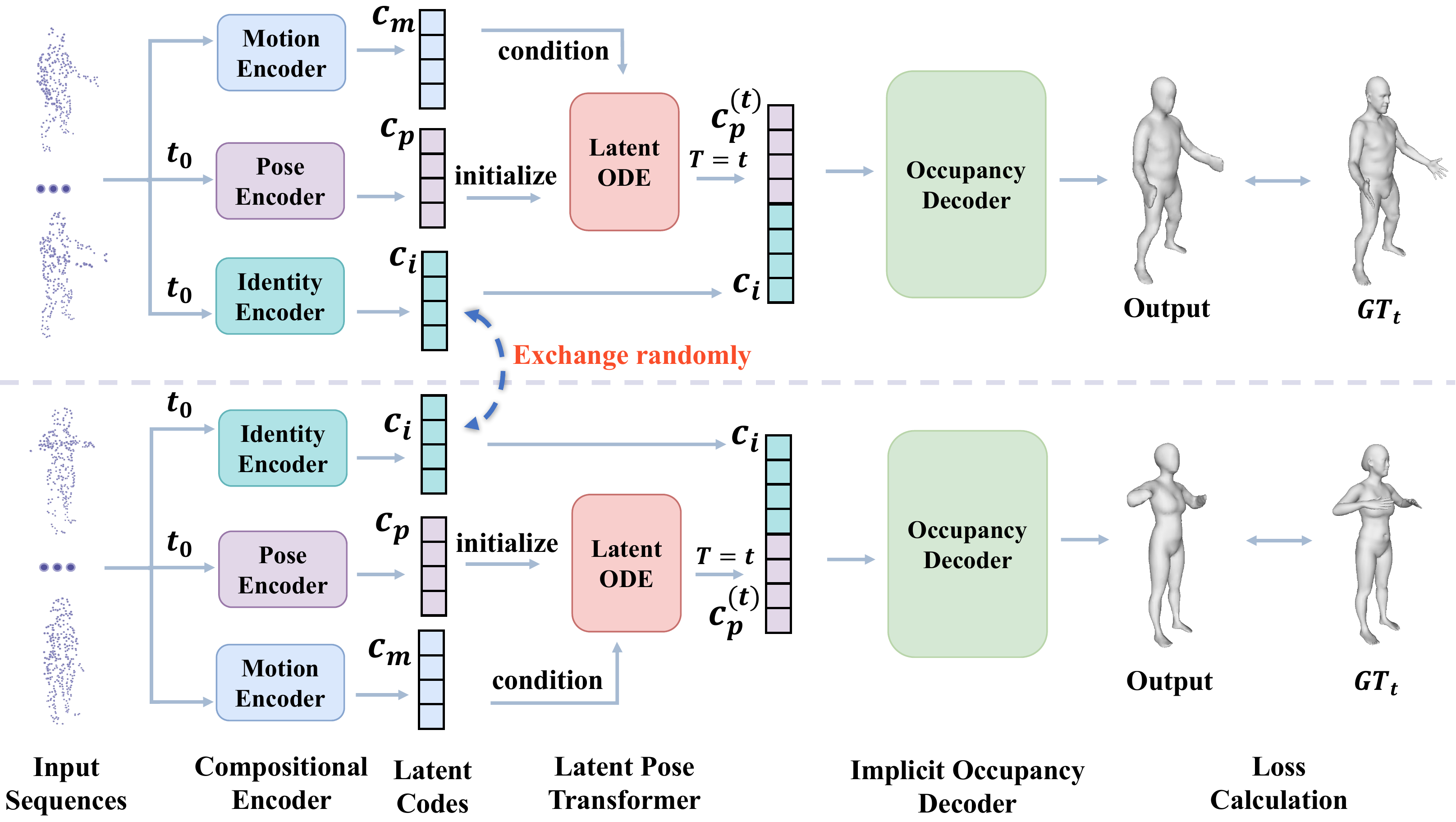}
\caption{\textbf{Overview of our model.} Our full model contains three building blocks, namely, compositional encoder, latent pose transformer and implicit occupancy decoder. During each training step, two point cloud sequences are chosen randomly from the training set as a pair and fed into the three encoders successively (note that the motion encoder is provided with the whole sequence, while the other two encoders are only provided with the first time step of sequence). After that, there is a 50\% probability that the identity codes of the two sequences are exchanged before continuing the forward propagation. Note that if the identity codes are exchanged, the ground truth meshes used for loss calculation will also be updated correspondingly.}
\vspace{-0.15in}
\label{fig:overview} 
\vspace{-0.1in}
\end{figure*}

 \section{Related Work}
There is a large body of work that focuses on 3D representation, 4D capture, 3D pose and motion transfer, and compositional/disentangled representation. We discuss the most related techniques in the context of our work.

\noindent \textbf{3D Representation} There has been a lot of work aiming at reconstructing a continuous surface from various type of inputs, such as color images~\cite{TatarchenkoDB17,pixel2mesh,KatoUH18, pixel2mesh++,Niu0018}, point clouds~\cite{boissonnat1984geometric,amenta1998new,kazhdan2006poisson}, etc. Recently, great success has been achieved for 3D shape reconstruction using deep learning techniques. In early works, 3D volumes~\cite{choy20163d,GirdharFRG16,wang2017cnn} and point clouds~\cite{qi2016pointnet,fan2017point,QiLWSG18,achlioptas2017representation} are adopted as the outputs of the networks, which suffer from the problems of losing surface details or limited resolutions. With the development of the graph convolution network, recent methods~\cite{groueix2018atlasnet,kanazawa2018learning,pixel2mesh,liao2018deep} take the triangle mesh as the output representation,  most of which regress the vertices and faces directly and require an initial template and fixed topology. Most recently, there has been significant work~\cite{Occupancy_Networks, park2019deepsdf,jiang2020local,chibane2020implicit,erler2020points2surf,chabra2020deep} on learning an implicit field function for surface representation, which allow more flexible output topology and network architectures. Among those methods, Occupancy Network~\cite{Occupancy_Networks} represents 3D shapes using continuous indicator functions by specifying which subset of 3D space the object occupies, and the iso-surface can be extracted by utilizing Marching Cube algorithm~\cite{LorensenC87}.

\noindent \textbf{4D Capture} Research on 4D capture has been advancing significantly in the past decades~\cite{pekelny2008articulated,ulusoy2014image,menze2015object,alldieck2018video}. However, most works are developed based on strong assumptions~\cite{wand2007reconstruction,pekelny2008articulated,ulusoy2014image,mehta2020xnect,wang2020sequential}, demand the costly multi-view inputs~\cite{neumann2002spatio,ulusoy2013dynamic,mustafa2015general,dong2020motion}. Behl \etal\cite{behl2017bounding} provide the 4D scene flow estimation leveraging object localization or semantic priors from deep networks, while the motion of scenarios is assumed to be in a tiny range, fixed pattern, rigid or linear, and high quality multi-view inputs are required. This greatly limits the ease of use and stability. Meanwhile, some methods exploit guided transformations on predefined templates to capture the time-dependent 3D flow~\cite{blanz1999morphable,loper2015smpl,zheng20174d,pishchulin2017building,kanazawa2019learning}. Such methods usually focus on specific shape categories, and the performance is restricted by the characteristic and the generalization ability of the template model.

Recently, Occupancy Flow~\cite{niemeyer2019occupancy} is presented to learn a temporally continuous field to model the motion of every point in space and time with Neural ODE \cite{chen2018neural} and the continuous implicit occupancy representation. Nevertheless, since the network is trained to model the continuous flow of the initial occupancy space, the quality of 4D reconstruction results relies on the initial frame heavily.

\noindent \textbf{3D Pose and Motion Transfer}
Conventional methods solving the 3D pose transfer problem via discrete deformation transfer. Learning-based mesh deformation is presented in \cite{wang2020neural}, which leverages the spatially adaptive instance normalization~\cite{huang2017arbitrary} in the network. Nevertheless, dense triangle mesh is required and the modeling of both spatial and temporal motion continuous flow is unavailable.

3D motion transfer aims at producing a new shape sequence given a pair of source and target shape sequences, making the target shape sequence do the same temporal deformation as the source, which focuses on the continuous pose transformation among shape sequences. By applying vector field-based motion code to target shape, Occupancy Flow~\cite{niemeyer2019occupancy} transfers motion among human model sequences. 
Essentially, since Occupancy Flow does not explicitly disentangle the representations of pose and shape as done in our work, we notice the good motion transfer results of Occupancy Flow are mostly achieved in the cases that the identities and initial poses from source and target are similar.

\noindent \textbf{Compositional/Disentangled Representation}
Learning compositional/disentangled representations has been extensively studied in previous work \cite{tokmakov2019learning,misra2017red,stone2017teaching,zhu2018visual,park2020swapping,tewari2019fml}. One attractive property of human intelligence is to learn novel concepts from a few or even a single example by composing known primitives \cite{tokmakov2019learning}, which is lacking in the current deep learning system. Prior work utilize compositional/disentangled representations to address various tasks. Zhu \etal \cite{zhu2018visual} disentangle shape, viewpoint, and texture and present an
end-to-end adversarial learning framework to generate realistic images. Tewari \etal \cite{tewari2019fml} learn a face model from in-the-wild video with a novel multi-frame consistency loss, the proposed approach represents
the facial geometry and appearance in different spaces and achieves realistic 3D face reconstruction. Park \etal \cite{park2020swapping} propose a fully unsupervised method to learn a swapping autoencoder for deep image manipulation task, which disentangles texture from structure. Most recently, Rempe \etal \cite{rempe2020caspr} propose CaSPR to learn a 4D representation of dynamic object point cloud sequences in Temporal-NOCS using latent Neural ODE and enable multiple applications. By dividing the latent feature into static and dynamic parts, it realizes shape and motion disentanglement. Unlike methods mentioned above, our goal is to learn a deep compositional representation for 4D captures with conditional latent Neural ODE, which decouples geometry template, initial state, and temporal deformation into different latent spaces, and supports various high-level tasks.


\section{Method}
In this section, we introduce our compositional representation for 4D captures and the training strategy to learn it from data.
The full pipeline of our framework is illustrated in Fig.~\ref{fig:overview}.
Taking a 3D model doing non-rigid deformation in time span $[0,1]$, we extract the sparse point cloud from the 3D model in $K$ uniformly sampled time stamps and feed them to the network as inputs.
Our goal is to learn separate compact representations for identity $\mathbf{c}_{i}$, initial pose $\mathbf{c}_{p}$, and motion $\mathbf{c}_{m}$, and reconstruct the 3D model at any continuous time stamp from them.
On the encoder side, we train three PointNet \cite{qi2016pointnet} based networks to extract $\mathbf{c}_{i}$ and $\mathbf{c}_{p}$ from the first frame, and $\mathbf{c}_{m}$ from the whole sequence.
To reconstruct mesh in target time $t$, we first update the initial pose code $\mathbf{c}_{p}$ to $\mathbf{c}_{p}^{\left(t\right)}$ encoding the pose of the 3D model in target time, which is achieved via a Neural ODE \cite{chen2018neural} conditioned on motion code $\mathbf{c}_{m}$.
The $\mathbf{c}_{i}$ and $\mathbf{c}_{p}^{\left(t\right)}$ are then concatenated and fed into a network to produce an implicit occupancy field indicating whether a 3D location is inside or outside the 3D shape, and the 3D mesh surface can be reconstructed via the Marching Cube algorithm \cite{lorensen1987marching}.

\subsection{Compositional Encoder}
We utilize three separate encoders to extract 128-d latent codes for identity, pose, and motion respectively. 
Inspired by Occupancy Flow \cite{niemeyer2019occupancy}, we use a PointNet-based~\cite{qi2016pointnet} network architecture with ResNet blocks~\cite{he2016deep} as the backbone, although the input to each encoder is different according to the semantic meaning of each code.
The initial pose only depends on the 3D shape in the first frame, therefore the corresponding encoder only takes the point cloud of the first frame,  i.e. $t=0$, as input.
In contrast, the motion encoder takes the whole point cloud sequence as the input since the motion code needs to encode the deformation throughout the whole time span.
To get the identity code, the encoder can take the whole sequence as input, but we empirically found that using only the first frame is enough and achieves similar performance.

\subsection{Latent Pose Transformer}
After obtaining the $\mathbf{c}_{i}$, $\mathbf{c}_p$, and $\mathbf{c}_{m}$ from the compositional encoder, the next step is to update the pose code for target time $t$, i.e. $\mathbf{c}_{p}^{\left(t\right)}$, which are used to reconstruct the 3D shape in corresponding time stamp.
Intuitively, the target pose code should start from the initial pose code, i.e. $\mathbf{c}_{p}^{\left(t=0\right)}=\mathbf{c}_p$, and varies continuously over time conditioned on the motion code $\mathbf{c}_{m}$.
To this end, we propose a novel latent pose transformer, which is achieved by a conditional latent Neural ODE.

Neural ODE is used to reconstruct continuous temporal signals $S(t)$.
Instead of directly estimating the target value, Neural ODE $f_\theta(t)$ predicts differential elements at each time stamp, which can be integrated to reconstruct the signal, i.e. $S(T) = S(0)+\int_{0}^{T}f_{\theta}\left(t,S(t)\right)dt$.
For our specific scenario, we train a Neural ODE to predict the variation of the latent pose code over time.
Different from the original Neural ODE, our model is further conditioned on the motion code, which allows the same network to update initial poses in different manners according to the motion exhibit in the input sequence.
Therefore, the pose code in target time $T$ is obtained by
\begin{equation}
\mathbf{c}_{p}^{\left(T\right)}=\mathbf{c}_{p}+\int_{0}^{T}f_{\theta}\left(\mathbf{c}_{p}^{\left(t\right)},t\mid\mathbf{c}_{m}\right)dt, \label{eq:LPT}
\end{equation}
where $f_{\theta}(\cdot)$ is modeled by a neural network
of 5 residual blocks with $\theta$ as the parameters.  
Following the advice of \cite{chen2018neural}, we obtain the gradient
using the adjoint sensitivity method \cite{pontryagin2018mathematical}.
For more details, please refer to the Supplementary Material.

\subsection{Implicit Occupancy Decoder}
The last stage of our model translates the identity code $\mathbf{c}_{i}$ and the pose code in target time $\mathbf{c}_{p}^{\left(t\right)}$ into 3D shape.
Inspired by high geometry quality from recent work using implicit representation \cite{park2019deepsdf,Occupancy_Networks,xu2019disn}, we train an Occupancy Network (ONet) \cite{Occupancy_Networks} to predict for each 3D location if they were inside or outside the object surface:
\begin{equation}
\mathbf{o}_{\mathbf{p}}^{\left(t\right)}:=\Phi_{\eta}\left(\mathbf{p}\mid\mathbf{c}_{i}\oplus\mathbf{c}_{p}^{\left(t\right)}\right),
\end{equation}
where $\mathbf{p}$ is a 3D location, $\Phi_{\eta}$ is an ONet parameterized by $\eta$, $\oplus$ denotes the concatenation operation between codes, $\mathbf{o}_{\mathbf{p}}^{\left(t\right)}$ is the occupancy of location $\mathbf{p}$ in time $t$.
Note that the identity code $\mathbf{c}_{i}$ remains the same as it should not change over time.


\subsection{Identity Exchange Training}{\label{sec:training}}
Naively training our network with 4D reconstruction is not sufficient to learn the compositional representation that isolates identity, initial pose, and motion.
We introduce a simple yet effective training strategy (shown in Fig.~\ref{fig:overview}), where the network is asked to reconstruct the same motion with different identities. 
Specifically, we extract latent codes for two sequences, $\left\{ \mathbf{c}_{i_{1}},\mathbf{c}_{m_{1}},\mathbf{c}_{p_{1}}\right\}$
and $\left\{ \mathbf{c}_{i_{2}},\mathbf{c}_{m_{2}},\mathbf{c}_{p_{2}}\right\}$, from different subjects $\mathbf{s}_{1}$ and $\mathbf{s}_{2}$.
We then swap their identity codes and supervise the model to reconstruct ground truth 4D captures of the same motion performed by the other subjects, i.e. $\mathbf{s}_{1}$ performing motion of $\mathbf{s}_{2}$ and vice versa.
Since $\mathbf{c}_{m_{1}}$ and $\mathbf{c}_{p_{1}}$ has no visibility to $\mathbf{s}_{2}$, all the identity information for $\mathbf{s}_{2}$ has to be encoded in $\mathbf{c}_{i_{2}}$ for successful reconstruction.
In practice, we perform this identity exchange training strategy for $50\%$ of the iterations, and find it effective in disentangling identity and motion.

\subsection{Loss Function}{\label{sec:loss_func}}
Our model is trained by minimizing the binary cross entropy error (BCE) on the occupancy of 3D locations.
Inspired by Occupancy Network \cite{Occupancy_Networks}, we randomly sample a time step $\tau$ and a set of 3D query points $\mathcal{S}$, and compute the loss $\mathcal{L}^{\left(\tau\right)}$
between the predicted occupancy value $\mathbf{o}_{\mathbf{p}}^{\left(\tau\right)}$
and the ground truth $\hat{\mathbf{o}}_{\mathbf{p}}^{\left(\tau\right)}$ :
\begin{equation}
\mathcal{L}^{\left(\tau\right)}=\sum_{\mathbf{p}\in\mathcal{S}}\textbf{BCE}\left(\hat{\mathbf{o}}_{\mathbf{p}}^{\left(\tau\right)},\mathbf{o}_{\mathbf{p}}^{\left(\tau\right)}\right).
\end{equation}

To get $\mathcal{S}$, we normalize all the meshes to $[-0.5, 0.5]$ with the global scale and translation calculated from the dataset, and sample 50\% points uniformly in a bounding volume and 50\% points near the surface of the mesh.
We find the definition of the bounding volume affects the training performance, and experiment with two ways: 1) a fixed volume with the length of 1; 2) a tight bounding volume around the mesh, in our experiments.

During training, we also supervise the predicted occupancy value at
time step $0$ to ensure a high quality initialization. Therefore, the complete loss function is defined as:
\begin{equation}
\mathcal{L}=\lambda_{1}\mathcal{L}^{\left(0\right)} + \lambda_{2}\mathcal{L}^{\left(\tau\right)},
\end{equation}
where $\lambda_{1}=\lambda_{2}=1.0$ in our experiment. We use Adam optimizer with the learning rate as $10^{-4}$. The model is trained with batch size of 16 on a single NVIDIA RTX 2080Ti GPU.

\begin{figure*}[tb]
\centering \includegraphics[width=0.9\linewidth]{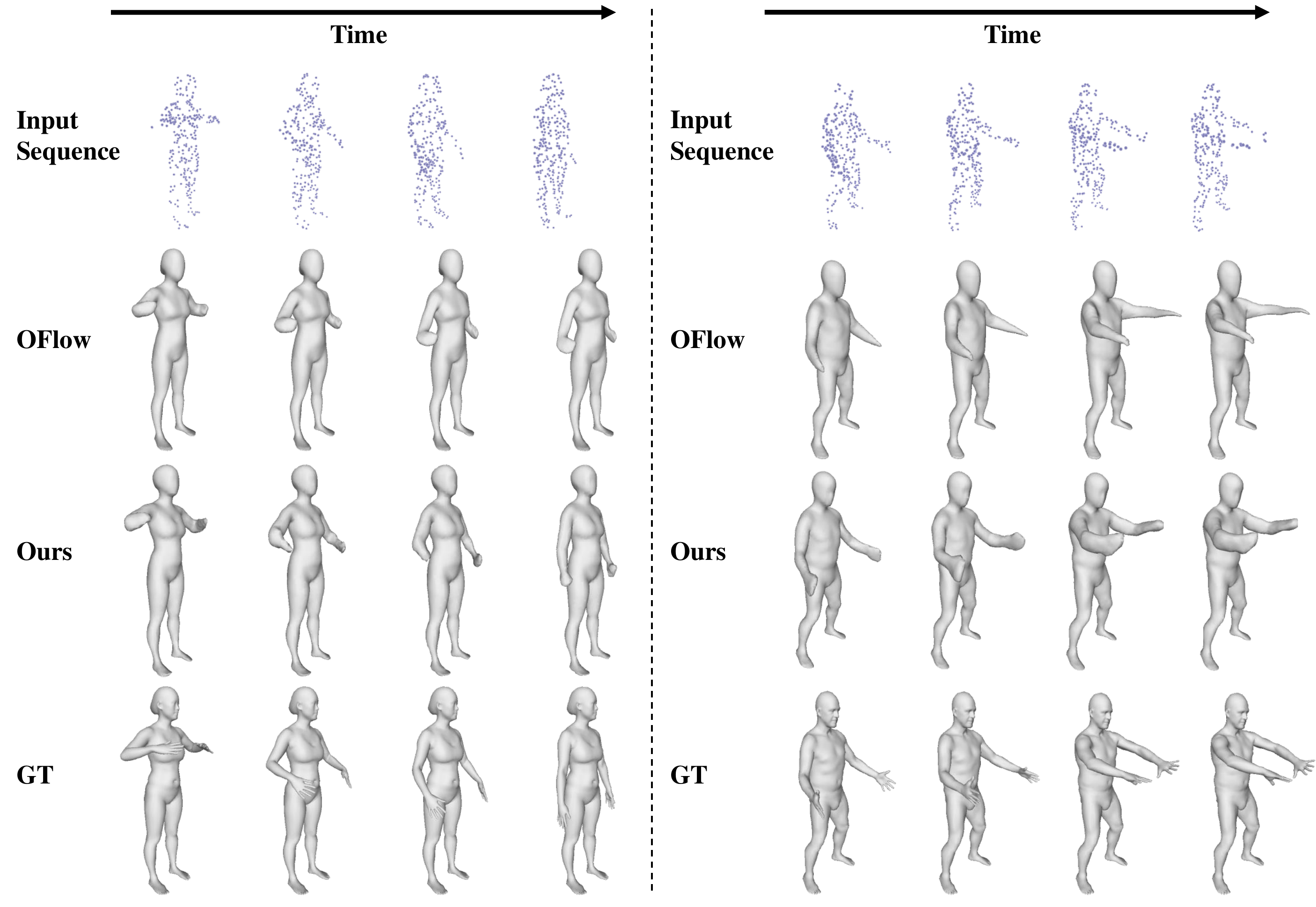}
\caption{\textbf{4D reconstruction from point cloud sequence (D-FAUST).} GT is short for Ground Truth.}
\label{fig:4d_recons} 
\vspace{-0.1in}
\end{figure*}

\begin{figure}[tb]
\centering \includegraphics[width=0.9\linewidth]{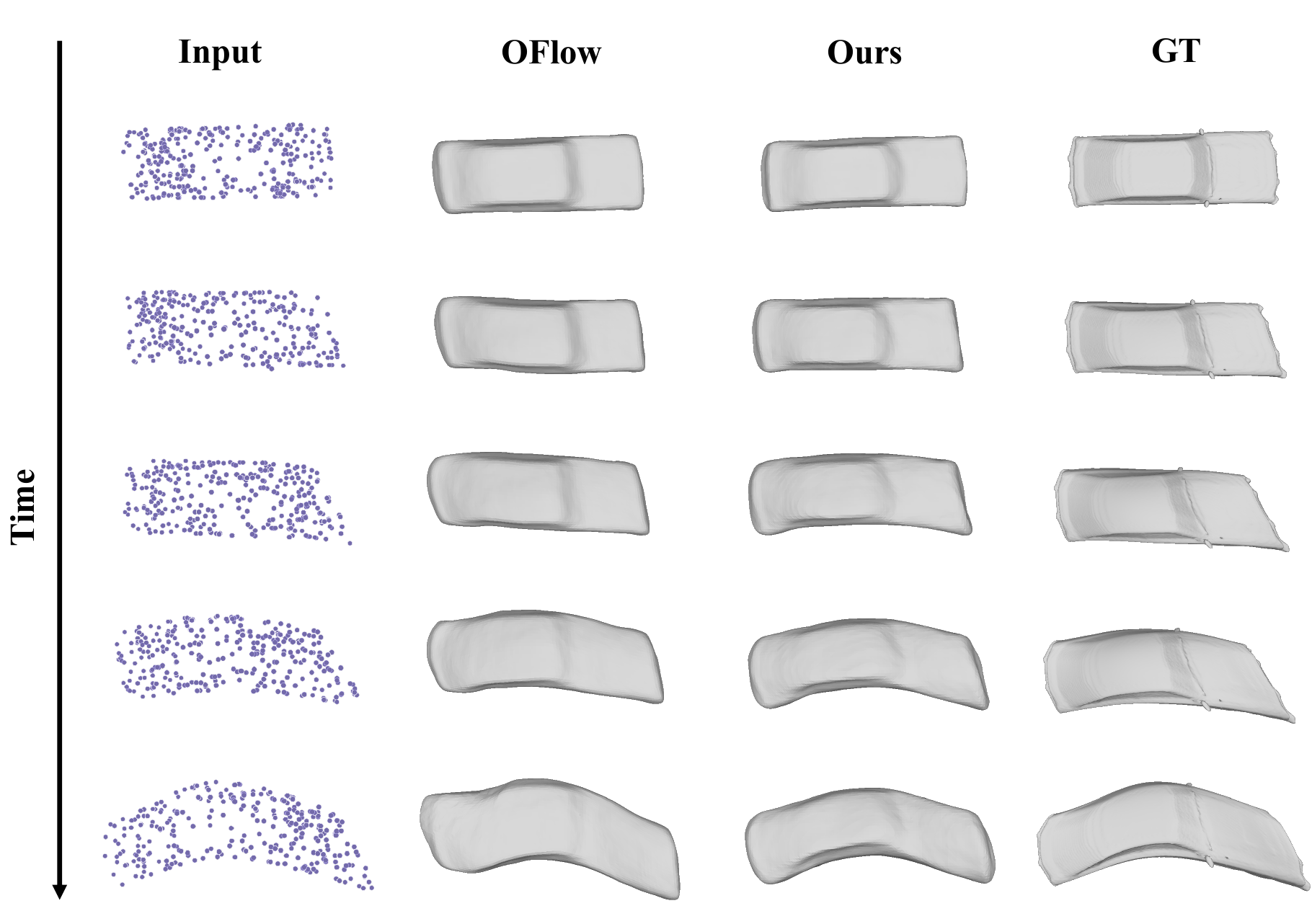}
\caption{\textbf{4D reconstruction on Warping Cars dataset.}}
\vspace{-0.1in}
\label{fig:4d_recons_cars} 
\vspace{-0.1in}
\end{figure}

\begin{table*}[tb]
\centering
 \begin{tabular}{lcccccc}
 \toprule[2px]
\multirow{2}{*}{Methods} & \multicolumn{2}{c}{Seen Individuals (D-FAUST)} & \multicolumn{2}{c}{Unseen individual (D-FAUST)} & \multicolumn{2}{c}{Warping Cars}\tabularnewline
\cline{2-7} \cline{3-7} \cline{4-7} \cline{5-7} \cline{6-7} \cline{7-7} 
 & IoU $\uparrow$ & Chamfer Distance $\downarrow$ & IoU $\uparrow$ & Chamfer Distance $\downarrow$ & IoU $\uparrow$ & Chamfer Distance $\downarrow$\tabularnewline
\toprule[2px]
PSGN 4D & - & 0.108 & - & 0.127 & - & -\tabularnewline
ONet 4D & 77.9\% & 0.084 & 66.6\% & 0.140 & - & - \tabularnewline
OFlow & 79.9\% & 0.073 & 69.6\% & 0.095 & 70.0\% & 0.166\tabularnewline
Ours & \textbf{81.8\%} & \textbf{0.068} & 68.2\% & 0.100 & \textbf{70.9\%} & \textbf{0.154} \tabularnewline
Ours* & 81.5\% & \textbf{0.068} & \textbf{69.9\%} & \textbf{0.094} & - & - \tabularnewline
\toprule[2px]
\end{tabular}
\vspace{-0.1in}
\caption{\label{tab:4dreconstruction}\textbf{4D reconstruction from point cloud sequence (D-FAUST and Warping Cars).} \textbf{Ours*} indicates that when preparing training data, we sample query points in a fixed cubic bounding volume with the length of 1 (see Sec. \ref{sec:loss_func}).}
\vspace{-0.1in}
\end{table*}

\section{Experiments}
In this section, we perform extensive experiments to evaluate our method. We first show the ability for 4D reconstruction, and then apply our compositional representation to various tasks like motion transfer, 4D completion and future prediction.

\subsection{Data Preparation}{\label{sec:data_preparation}}
We use two datasets to train and evaluate our proposed method.
The first dataset is Dynamic FAUST (D-FAUST) \cite{bogo2017dynamic}, which contains 129 mesh sequences of 10 real human subjects performing 14 different motions and all meshes are registered with SMPL \cite{loper2015smpl} model.
We augment D-FAUST to meet the needs of our Identity Exchange Training strategy (Sec. \ref{sec:training}).
We first fit the SMPL shape and pose parameters for all the data.
Then, the ground truth mesh sequences of all the combinations of identities and motions are generated, extending the number of mesh sequences to about 1000.

We also build a Warping Cars dataset using the approach introduced in Occupancy Flow \cite{niemeyer2019occupancy} to investigate the performance of our method on non-human objects. Specifically, we randomly choose 10 car models from ShapeNet \cite{shapenet2015} Car category and generate 1000 warpings. To generate a warping field, Gaussian displacement vectors are sampled in a $3\times3\times3\times5$ grid and the RBF \cite{park1991universal} interpolation is used to obtain a continuous displacement field. We combine different car shapes and warpings and finally get the dataset with total number of mesh sequences to 10000, each of which has 50 time steps.

\subsection{4D Reconstruction}
We first verify the reconstruction ability of our model following the setting in Occupancy Flow (OFlow)~\cite{niemeyer2019occupancy}.
The network consumes 300 sparse point trajectories as input, each of which consists of 3D locations at $L=17$ equally divided time stamps, and the goal is to reconstruct compact mesh at these time stamps even though the model is able to produce mesh at any particular time.
For human model, we use the same train/test split on D-FAUST \cite{bogo2017dynamic} as OFlow, including data on subjects seen and unseen during the training respectively.
For Warping Cars dataset, we test on our own testing set as it was not released.

The quantitative results on the D-FAUST dataset and the Warping Cars dataset are summarized in Tab.~\ref{tab:4dreconstruction}, where we report the average IoU and Chamfer Distance over 17 frames of all testing sequences.
As our baseline, ``PSGN 4D'' is a 4D extension of Point Set Generation Network \cite{fan2017point} by predicting a set of trajectories instead of single points, and ``ONet 4D'' is an extension of Occupancy Network (ONet) \cite{Occupancy_Networks}, which predicts occupancy value for points sampled in 4D space and reconstructs each frame of the sequence separately. 
OFlow uses Neural ODE to learn a continuous motion vector field for every point in space and time. While OFlow explicitly transform the 3D coordinates of each point,  we transform the pose code in the latent space. 
The results of baselines on the D-FAUST dataset are cited from OFlow \cite{niemeyer2019occupancy}, and the results on Warping Cars dataset are produced by a retrained OFlow.
Overall, our method performs comparable or better than other methods on D-FAUST and Warping Cars datasets, indicating that our model is able to reconstruct accurate surfaces. 

In Fig.~\ref{fig:4d_recons}, we show qualitative comparison on the D-FAUST dataset with OFlow. Our method is able to capture more details, such as the shape of the opening hands and the outline of the muscles on the body. In particular, OFlow fails to track the motion of hands in the last frame of the left sequence, while our method produces stable results during the whole sequence time.
This is presumably because our method reconstructs each frame of the whole sequence individually with the transformed pose latent code, while OFlow only reconstructs the first frame and deforms it with the learned transformation flow. Furthermore, the qualitative results on the Warping Cars dataset are shown in Fig.~\ref{fig:4d_recons_cars}, in which our method shows better capability of recovering motion than OFlow.

We provide an ablation study on 4D reconstruction task based on the D-FAUST dataset to evaluate the effectiveness of ODE in different aspects.
\textit{1) ODE in feature v.s. 3D.} We found that training an ODE directly in 3D space, i.e. exactly an OFlow, with identity exchange is hard to converge. We trained the model for a week on a single NVIDIA RTX 2080Ti GPU, and found the model learns barely any motion. In contrast, applying ODE in feature space, i.e. our method, may benefit from regularization provided by the compact 1-d latent vector and converges well. \textit{2) ODE v.s. MLP.} We train a model replacing ODE to an MLP that directly produces the pose code for a specified time, and get IoU=$80.4\%$ and Chamfer Distance=0.073 for reconstruction on D-FAUST, while our ODE model achieves IoU=$81.8\%$ and Chamfer Distance=0.068. This indicates that ODE performs better than MLP in reconstructing pose sequence for motion, but the MLP model trains and runs relatively faster.


\subsection{Pose and Motion Transfer}
Our compositional representation also naturally supports motion transfer.
Consider two subjects performing different motions, namely ${id_1+motion_1}$ and ${id_2+motion_2}$, and our goal is to generate 4D sequence with ${id_2+motion_1}$.
To do so, we first extract the latent representations with our compositional encoder for each input sequence, namely $\left(\mathbf{c}_{i}^{1},\mathbf{c}_{p_{0}}^{1},\mathbf{c}_{m}^{1}\right),\left(\mathbf{c}_{i}^{2},\mathbf{c}_{p_{0}}^{2},\mathbf{c}_{m}^{2}\right)$, and then feed $\left(\mathbf{c}_{i}^{2}, \mathbf{c}_{p_{0}}^{1},\mathbf{c}_{m}^{1}\right)$ to the latent pose transformer and implicit occupancy decoder.

We evaluate our method on the D-FAUST testing set, where we randomly select 20 identity and motion pairs, and generate the ground truth 4D sequences after motion transfer using the known SMPL parameters.
As baseline, we compare to OFlow which also learns separate codes to represent first frame geometry (i.e. the identity) and velocity field (i.e. the motion) respectively.
In addition, we also build a baseline with the recent state-of-the-art neural pose transfer method NPT~\cite{wang2020neural}, which utilizes spatially adaptive instance normalization to deform the identity point cloud to each time step of the target motion sequence using pose transfer.
The transformed point clouds are then fed into OFlow to generate complete meshes.

\begin{figure}[ht]
\centering \includegraphics[width=0.95\linewidth]{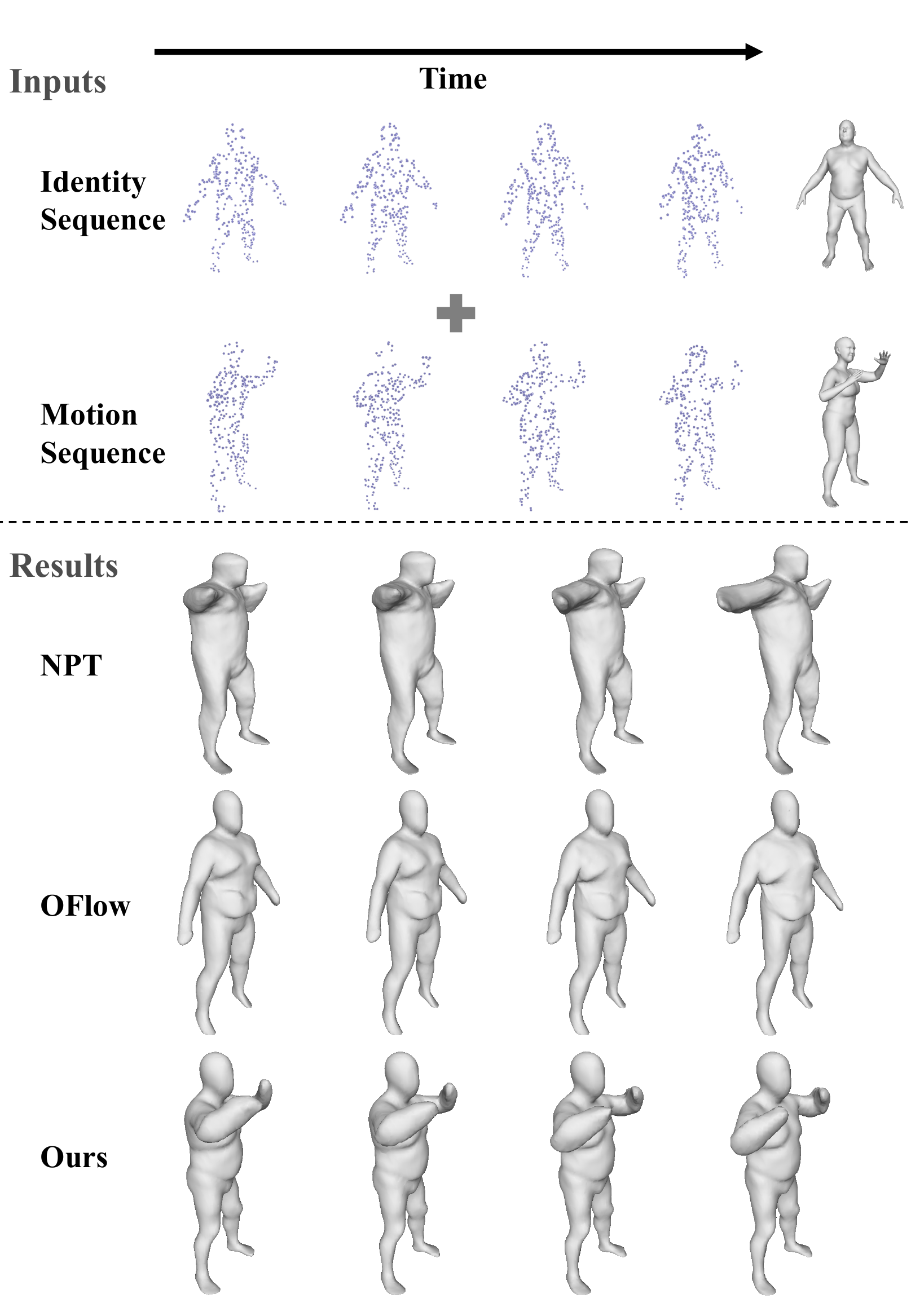}
\vspace{-0.1in}
\caption{\textbf{Motion transfer results on an input sequence pair (D-FAUST).} Our method transfers the motion of the second sequence to the identity of the first sequence successfully, while keeping the shape property of the first sequence unchanged.}
\label{fig:motion_transfer} 
\vspace{-0.1in}
\end{figure}

The quantitative results are shown in Tab.~\ref{tab:motion_transfer}.
Our method significantly outperforms other baseline methods with large margins.
One qualitative comparison is shown in Fig.~\ref{fig:motion_transfer}.
The performance of NPT is heavily limited by the density of the input identity and motion sequences, which makes it hard to transfer the continuous motion with sparse inputs.
OFlow does not transfer the motion at all, presumably because the pose representations are not decoupled from the shape latent code, which leads to a wrong first frame pose initialization and the failure of the whole generated sequence.
In contrast, our method successfully transfers the motion to the new identity, including both the initial pose and following frames.
Additional results on Warping Cars dataset are shown in Supplementary Material.

\vspace{-0.1in}
\begin{table}[htb]
    \centering
    \begin{tabular}{lcc}
     \toprule[2px]
     & IoU $\uparrow$ & Chamfer Distance $\downarrow$\\
     \toprule[2px]
    NPT & 26.4\% &  0.498\\
    OFlow & 26.7\% &  0.400\\
    Ours & \textbf{85.0\%} &  \textbf{0.055} \\
     \toprule[2px]
    \end{tabular}
    \vspace{-0.1in}
\caption{\label{tab:motion_transfer}\textbf{Motion transfer (D-FAUST).}}
\vspace{-0.15in}
\end{table}

We further investigate if the motion code $\mathbf{c}_{m}$ can be transferred without the initial pose code $\mathbf{c}_{p}$.
Even though this is sometimes an ill-posed problem (e.g. forcing a stand-up motion to start with a standing pose), we find, surprisingly, our model is still able to produce reasonable results if the new initial pose is not too different from the original one (See Supplementary Material for results).
This indicates that our conditional Neural ODE is robust to some extent against the noise in the initial pose code.

\begin{table}[tb]
    \centering
\begin{tabular}{lcccc}
 \toprule[2px]
\multirow{2}{*}{Methods} & \multicolumn{2}{c}{Temporal} & \multicolumn{2}{c}{Spatial}\tabularnewline
\cline{2-5} \cline{3-5} \cline{4-5} \cline{5-5} 
 & IoU $\uparrow$ & CD $\downarrow$ & IoU $\uparrow$ & CD $\downarrow$\tabularnewline
 \toprule[2px]
OFlow & 85.1\% & 0.057 & 86.0\%  & 0.054\tabularnewline
Ours & 86.4\% & 0.056 & 87.2\% & 0.051\tabularnewline
 \toprule[2px]
\end{tabular}
\vspace{-0.1in}
\caption{\label{tab:4d_completion}\textbf{4D temporal completion and spatial completion (D-FAUST).} CD is short for Chamfer Distance.}
\vspace{-0.15in}
\end{table}

\begin{figure}[tb]
\centering \includegraphics[width=0.9\linewidth]{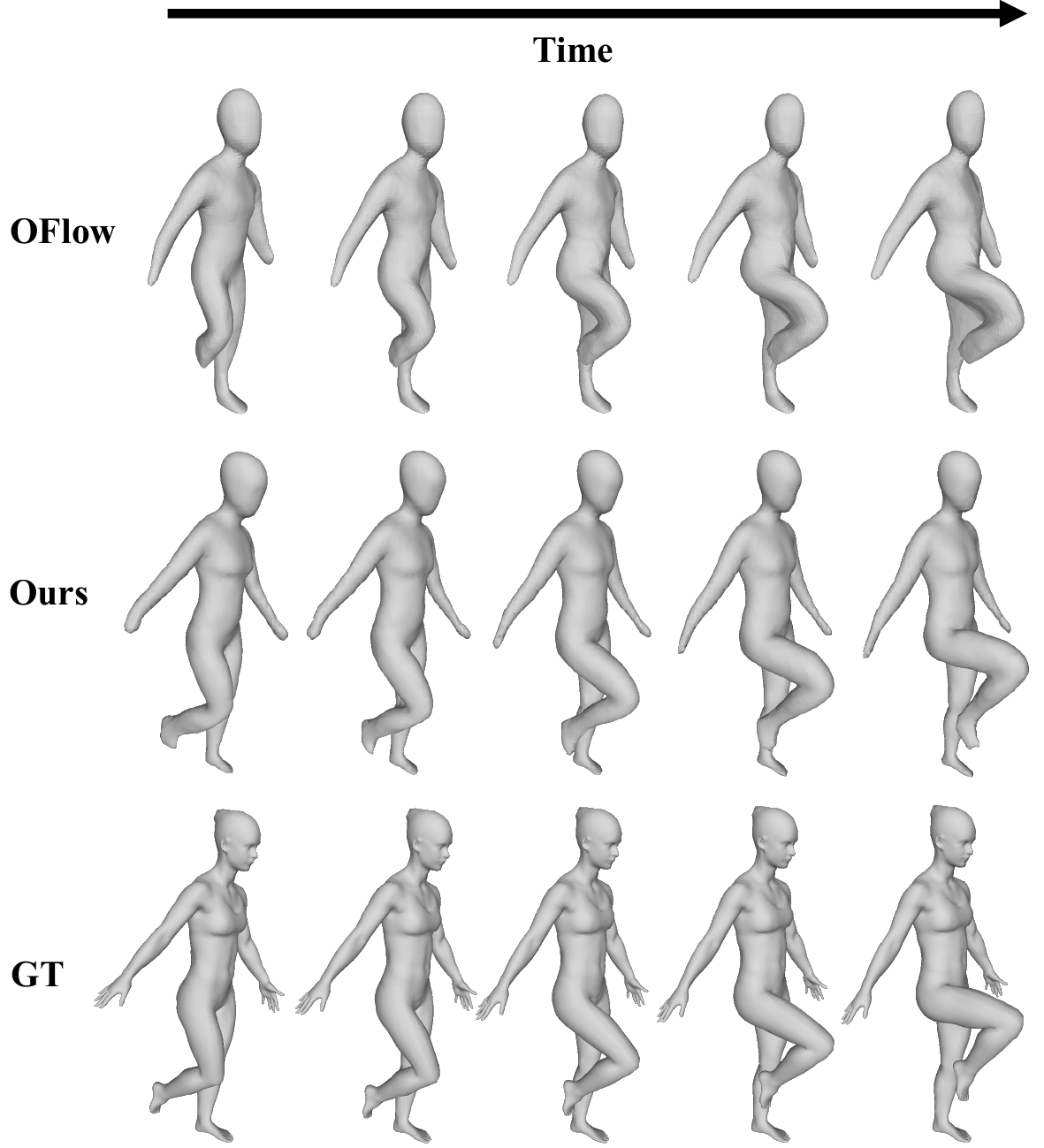}
\caption{\textbf{4D temporal completion (D-FAUST).} We show the results of 5 missing frames.}
\label{fig:temporal_completion} 
\vspace{-0.1in}
\end{figure}


\begin{figure}[tb]
\centering \includegraphics[width=0.9\linewidth]{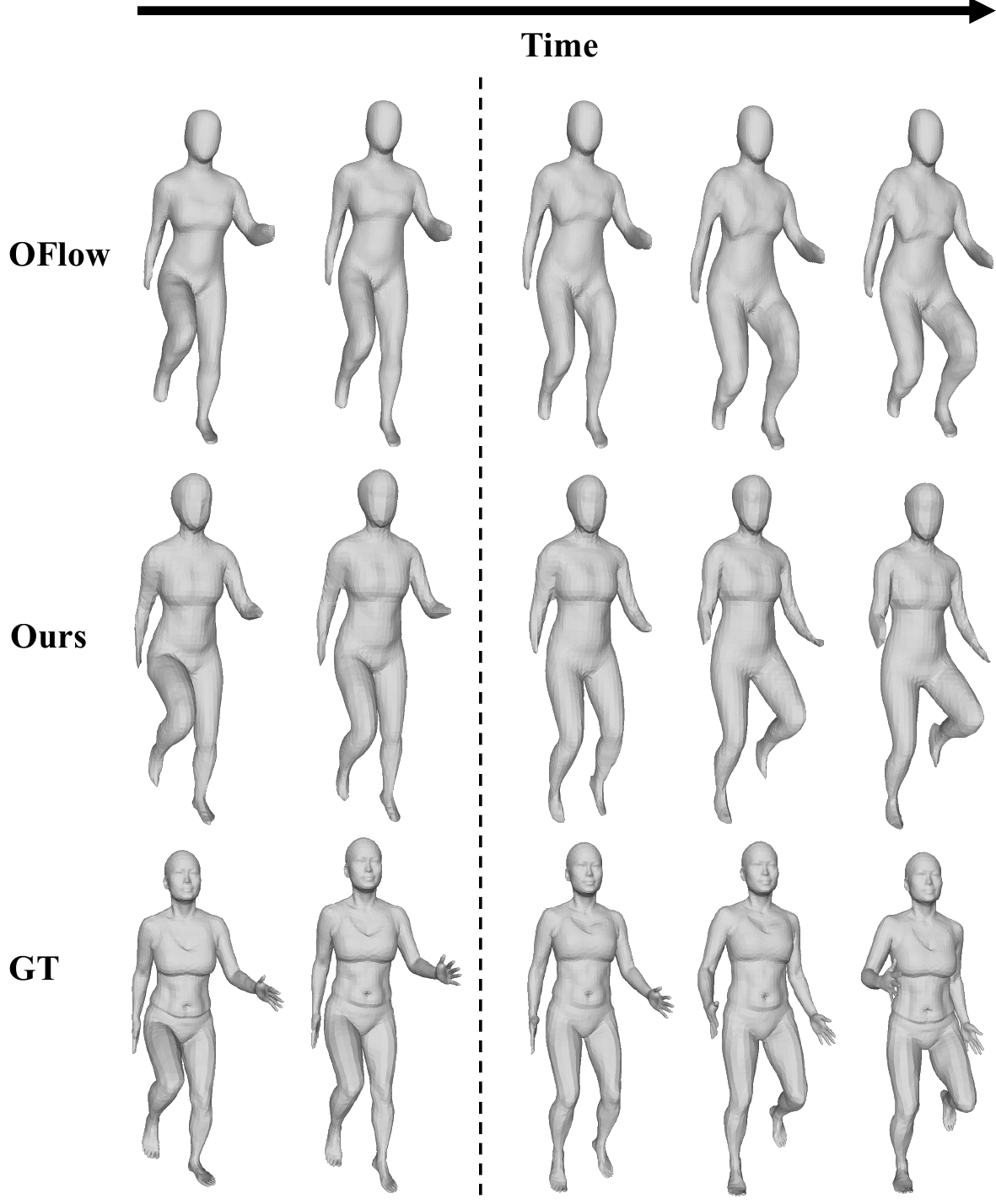}
\caption{\textbf{Future prediction (D-FAUST).} The results on the left of the dotted line are reconstructions for partial observation, and the results on the right are future predictions.}
\label{fig:future_pred} 
\vspace{-0.15in}
\end{figure}

\subsection{4D Completion}
Our compositional 4D representation also provides strong prior as the regularization for 4D completion task, in which the goal is to fill in the missing signals in a given 4D capture with only partial observation.
This is practically useful when part of the 4D capture is corrupted due to imperfect capturing techniques or challenging scenarios.
Specifically, this task can be categorized into two kinds based on the missing data: 1) \textbf{Temporal completion}, which recovers the missing frames; 2) \textbf{Spatial completion}, which completes partial geometry in each frame.
To perform these tasks, we remove the encoder, fix the decoder parameters, and optimize the latent codes with back-propagation until the output 4D sequence matches the partial observation.

The experiments are conducted on the D-FAUST dataset. For temporal completion, we select 18 mesh sequences with $L=30$ frames from the testing set, and randomly withheld half of the frames in each sequence for testing.
For spatial completion, we randomly select three points in each frame and remove the points less than 0.2 away from them.

Comparison to OFlow is shown in Tab.~\ref{tab:4d_completion}.
Our method performs comparable or better IoU and Chamfer distance than OFlow on both temporal and spatial completion.
Fig. \ref{fig:temporal_completion} shows a temporal completion result.
Our method successfully interpolates correct poses for missing frames with more complete geometry than OFlow.
Please refer to Supplementary Material for results on Warping Cars dataset.

\subsection{Future Prediction}
Not only interpolating internal missing frames, our model can also predict the future of the motion by extrapolating temporal frames onward.
To validate this, we select 15 mesh sequences with $L=20$ frames from the D-FAUST testing set, and always remove the last 10 frames instead of randomly selected ones.
Tab. \ref{tab:future_pred} and Fig. \ref{fig:future_pred} show the comparison to OFlow.
Though OFlow can also produce reasonable future motion, the magnitudes are usually small which leads to overly slow motion.
In contrast, our method predicts much more accurate motion, e.g. with the other leg raised. The results on Warping Cars dataset are shown in Supplementary Material.

\begin{table}[htb]
    \centering
    \begin{tabular}{lcc}
     \toprule[2px]
     & IoU $\uparrow$ & Chamfer Distance $\downarrow$\\
     \toprule[2px]
    OFlow & 75.5\% &  0.099\\
    Ours & 80.8\% &  0.081 \\
     \toprule[2px]
    \end{tabular}
    \vspace{-0.1in}
\caption{\label{tab:future_pred}\textbf{Future prediction (D-FAUST).} We remove the last 10 frames of sequence to investigate the extrapolation ability of our method.}
\vspace{-0.15in}
\end{table}

\section{Conclusion}

This paper introduces a compositional representation for 4D captures by disentangling the geometry template, initial state and temporal deformation with separated compact latent codes, which can reconstruct the deforming 3D object over a temporal span. Furthermore, an identity exchange training strategy is proposed to make geometry template and temporal deformation efficiently decoupled and exchangeable. Extensive experiments on 4D reconstruction, pose and motion transfer, 4D completion, and motion prediction validate the efficacy of our proposed approach.

\section*{Acknowledgement}
Yanwei Fu is the corresponding author. This work was supported in part by NSFC Projects (U62076067), Science and Technology Commission of Shanghai Municipality Projects (19511120700, 19ZR1471800), Shanghai Research and Innovation Functional Program (17DZ2260900), Shanghai Municipal Science and Technology Major Project (2018SHZDZX01) and ZJLab.

{\small
\bibliographystyle{ieee_fullname}
\bibliography{egbib}
}

\clearpage
\noindent \textbf{\Large Supplementary Material}
\setcounter{section}{0}

\begin{figure*}[tb]
\centering 

\subfloat[Encoder architecture]{       \centering    
\label{fig:encoder}           
\includegraphics[width=0.95\linewidth]{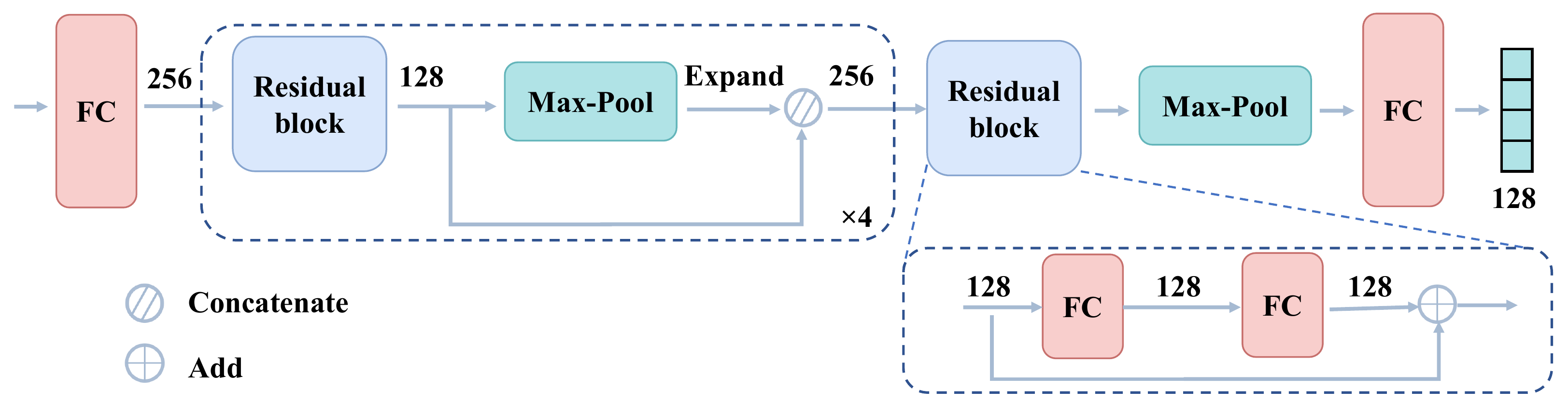}
}\\
\subfloat[Vector field network in our Latent Pose Transformer]{                  
    \label{fig:vector_field}            
\includegraphics[width=0.485\linewidth]{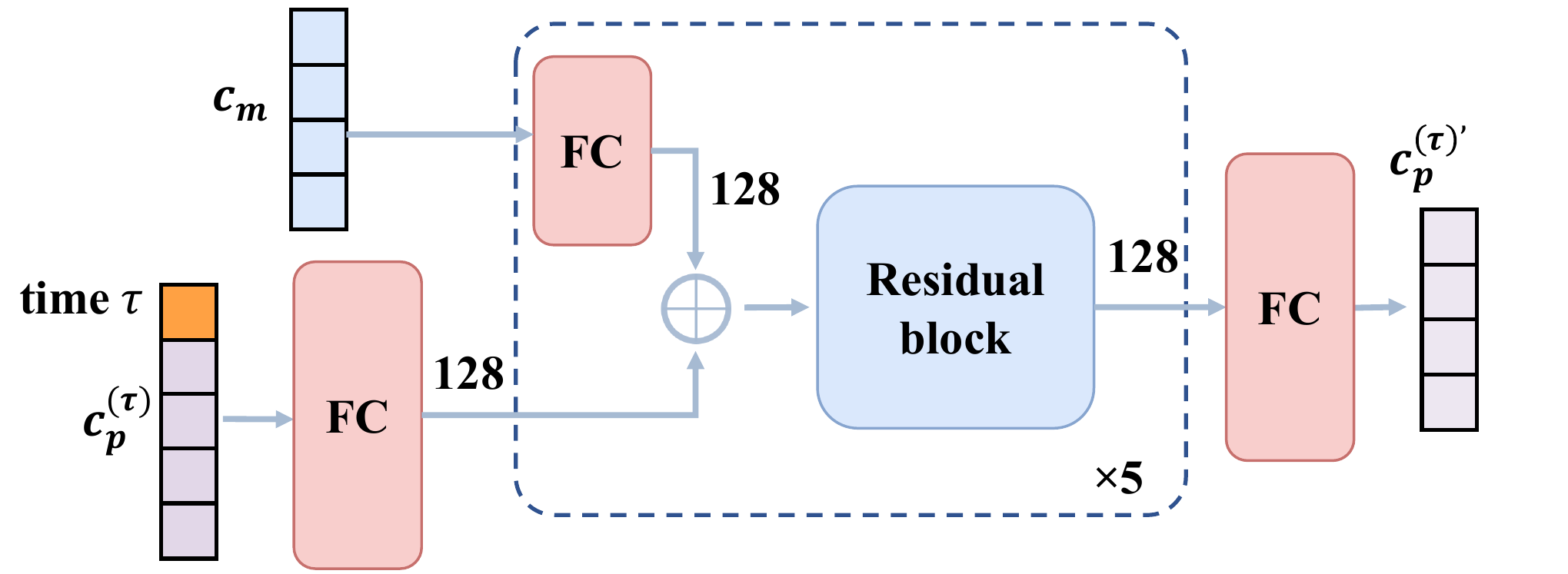}
}
\subfloat[Decoder architecture]{                    
    \label{fig:decoder}            
\includegraphics[width=0.485\linewidth]{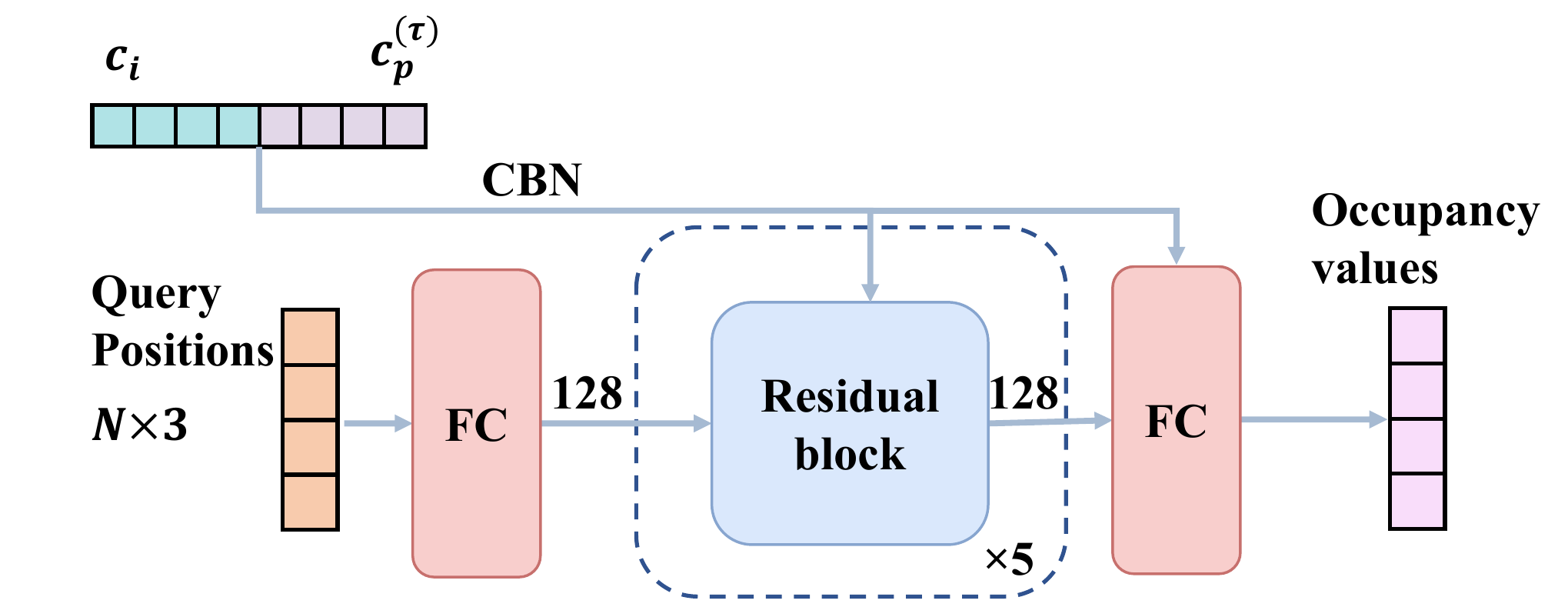}
}
\caption{\textbf{Detailed architectures of our framework.}}
\vspace{-0.1in}
\label{fig:architecture} 
\end{figure*}

\section{Implementation Details}{\label{imple_detail}}
In this section, we provide network architectures used for the compositional encoder, latent pose transformer, and implicit occupancy decoder in our framework. Additionally, we discuss more details about training and 4D completion experiment.

\subsection{Network Architecture}
\noindent \textbf{Compositional Encoder} The input of our encoder network is a point cloud sequence of size $\left(B,L,N,3\right)$, where $B,L,N$ denote batch size, length of input sequence and the number of points in each point cloud, respectively. The first frame of the point clouds is consumed by the identity encoder and pose encoder. For motion encoder, we concatenate
all the input point clouds along the last dimension and set the input dimension of our encoder network to $3L$. The encoder network is a variation of PointNet \cite{qi2016pointnet} which has five residual blocks as shown in Fig. \ref{fig:encoder}. Each of the first four blocks has an additional max-pooling operation to obtain aggregated feature of size $\left(B,1,C\right)$ where $C$ denotes the dimension of hidden layers, and an expansion operation (expand the pooled feature to the size $\left(B,N,C\right)$) to make it suitable for concatenation. The output of the fifth block is passed through a max-pooling layer and a fully connected layer to get the final latent vector of dimension 128.

\noindent \textbf{Latent Pose Transformer} Our latent pose transformer (LPT) is built as a latent ODE conditioned by the motion code, which contains a vector field network and the architecture is shown in Fig. \ref{fig:vector_field}. The vector field network is fed with the motion code and the concatenation of a time value $\tau$ and its corresponding pose code $c^{\left(\tau\right)}_{p}$ as inputs, then outputs the differential of pose code at time $\tau$. There are five residual blocks in the vector field network, and the input of each block is summed up with the feature encoded from the motion code. 

Given the motion code $c_m$, initial pose code $c_p$ and a queried time value $t$, our LPT evaluates the vector field network multiple times to obtain transformed pose code at time $t$, which has the same dimension as the initial pose code. 

\noindent \textbf{Implicit Occupancy Decoder} We utilize Occupancy Network (ONet) \cite{Occupancy_Networks} as our decoder (Fig. \ref{fig:decoder}), which
has the similar architecture with the vector field network. The decoder
gets a 3D query point from a set of sample points $\ensuremath{\mathcal{S}}$
and a conditioning code as input, and outputs a scalar value which
indicates the probability that the queried point is inside object surface. In our
framework, the conditioning code is the concatenation of the identity
code $\mathbf{c}_{i}$ and the pose code $\mathbf{c}_{p}^{\left(\tau\right)}$ at
time $\tau$.
Following ONet, we use the conditional batch normalization (CBN)
scheme to insert guidance encoded from the concatenated conditioning code.

\subsection{More Details and Hyper-parameters}
\noindent \textbf{Training} Our framework is implemented in PyTorch. For training, we use the Adam optimizer\cite{kingma2014adam} with the learning rate $10^{-4}$. The threshold for the output occupancy probability is set to 0.4.
We use the adaptive-step solver $\textit{dopri5}$ \cite{dormand1980family} with relative tolerance of $10^{-3}$ and absolute tolerance of $10^{-5}$. During training, 2048 points are sampled both at the initial time step $t=0$ and a randomly selected time step $t>0$ for every sequence to compute loss.

\noindent \textbf{4D Completion} For 4D completion, we use the same hyper-parameters for Occupancy Flow (OFlow) \cite{niemeyer2019occupancy} and our method. We initialize the latent codes to be Gaussian noise with standard deviation 0.1 and use the Adam optimizer with initial learning rate 0.03 to perform back-propagation for 500 iterations. The learning rate is decreased by half every 100 iterations. We reconstruct and compute BCE loss on all the observations in each iteration.

\section{Data Processing}{\label{data_detail}}
\noindent \textbf{D-FAUST}
For our Identity Exchange Training (IET) strategy, all combinations of human identities and motions are required. Since all the mesh models in the original D-FAUST dataset have registered with the SMPL \cite{loper2015smpl} model, we retrieve the SMPL identity and pose parameters for every mesh model by optimizing with back-propagation. The mean L2 distance between the predicted vertices and the ground truth vertices is used as the loss function.

We need point cloud sequences and query points for training purpose. When sampling the input point clouds, we do not perform a separate normalization for each model like OFlow. Instead, we keep the locations and scales of the original outputs of the SMPL model as they are already aligned. For sampling query points, we perform a global normalization for all the mesh models in our augmented dataset as described in Section 3.5 of the main paper.

\noindent \textbf{Warping Cars} We thank the authors of OFlow \cite{niemeyer2019occupancy} for sharing the code, and follow their paper to generate the Warping car dataset.
We choose 10 different car shape models in the watertight version of ShapeNet \cite{shapenet2015} ``Car'' category and generate 1000 warpings with the approach explained in Section 4.1 of the main paper.
We adopt the same strategy as OFlow to obtain the input point clouds (normalized to a unit cube) and query points (sample uniformly in the bound volume), because the mesh models in the ShapeNet are consistently aligned and scaled.

\section{Additional Ablation Study}{\label{ablation}}
\noindent \textbf{Impact of the Identity Exchange Rate} We train a set of models with the identity exchange rates set to $0\%, 25\%, 50\%, 75\%, 100\%$ respectively, and show the 4D reconstruction and motion transfer performance on D-FAUST dataset in Tab. \ref{tab:rate}.
The overall performances for both tasks are in general stable w.r.t. the exchange rate.
Though 4D reconstruction achieves the best accuracy at $0\%$, the model loses the shape/motion disentanglement and thus fails for motion transfer.
In general, with $50\%$, the model achieves the best motion transfer performance and reasonably high reconstruction accuracy.

\begin{table}[htb]
    \centering
\begin{tabular}{ccccc}
 \toprule[2px]
\multirow{2}{*}{Exchange Rate} & \multicolumn{2}{c}{4D Reconstruction} & \multicolumn{2}{c}{Motion Transfer}\tabularnewline
\cline{2-5} \cline{3-5} \cline{4-5} \cline{5-5} 
 & IoU $\uparrow$ & CD $\downarrow$ & IoU $\uparrow$ & CD $\downarrow$\tabularnewline
 \toprule[2px]
0\% & \textbf{83.3\%} & \textbf{0.061} & 65.3\%  & 0.137\tabularnewline
25\% & 81.8\% & 0.066 & 84.1\% & 0.057\tabularnewline
50\%$^\ast$ & 81.8\% & 0.068 & \textbf{85.0\%} & \textbf{0.055}\tabularnewline
75\% & 81.2\% & 0.068 & 83.7\% & 0.059\tabularnewline
100\% & 81.0\% & 0.070 & 84.4\% & 0.058\tabularnewline
 \toprule[2px]
\end{tabular}
\caption{\label{tab:rate}\textbf{Results about different choices of  the identity exchange rate during training.} $^\ast$ denotes our choice in the main paper. CD is short for Chamfer Distance.}
\end{table}

\begin{table*}[htb]
    \centering
\begin{tabular}{lcccccccc}
 \toprule[2px]
\multirow{2}{*}{Methods} & \multicolumn{2}{c}{Motion Transfer} & \multicolumn{2}{c}{Temporal Completion} & \multicolumn{2}{c}{Spatial Completion} & \multicolumn{2}{c}{Future Prediction}\tabularnewline
\cline{2-9} \cline{3-9} \cline{4-9} \cline{5-9} \cline{6-9} \cline{7-9} \cline{8-9} \cline{9-9} 
 & IoU $\uparrow$ & CD $\downarrow$ & IoU $\uparrow$ & CD $\downarrow$ & IoU $\uparrow$ & CD $\downarrow$ & IoU $\uparrow$ & CD $\downarrow$\tabularnewline
 \toprule[2px]
OFlow & 30.8\% & 0.596 & 78.8\% & 0.138 & 80.2\% & 0.130 & 57.6\% & 0.293\tabularnewline
Ours & 68.9\% & 0.181 & 81.6\% & 0.117 & 81.3\% & 0.121 & 63.6\% & 0.227\tabularnewline
 \toprule[2px]
\end{tabular}
\caption{\label{tab:various_car}\textbf{Comparisons to OFlow on various tasks for our generated Warping Cars dataset.}}
\end{table*}

\section{Various Tasks for Warping Cars}{\label{various_car}}

\begin{figure}[tb]
\centering \includegraphics[width=1\linewidth]{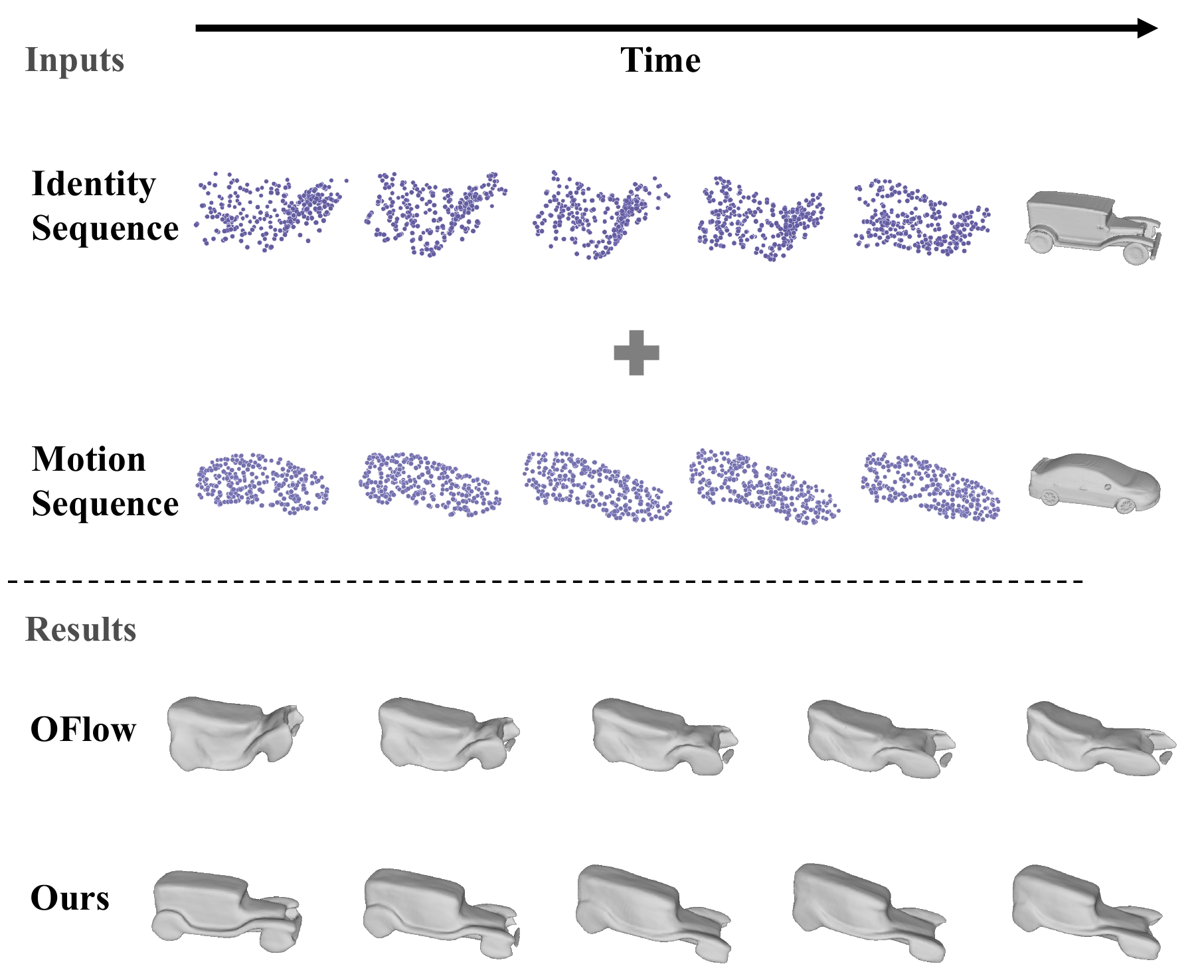}
\caption{\textbf{Motion transfer (Warping Cars).}}
\label{fig:transfer_car}
\end{figure}

\begin{figure}[tb]
\centering \includegraphics[width=0.85\linewidth]{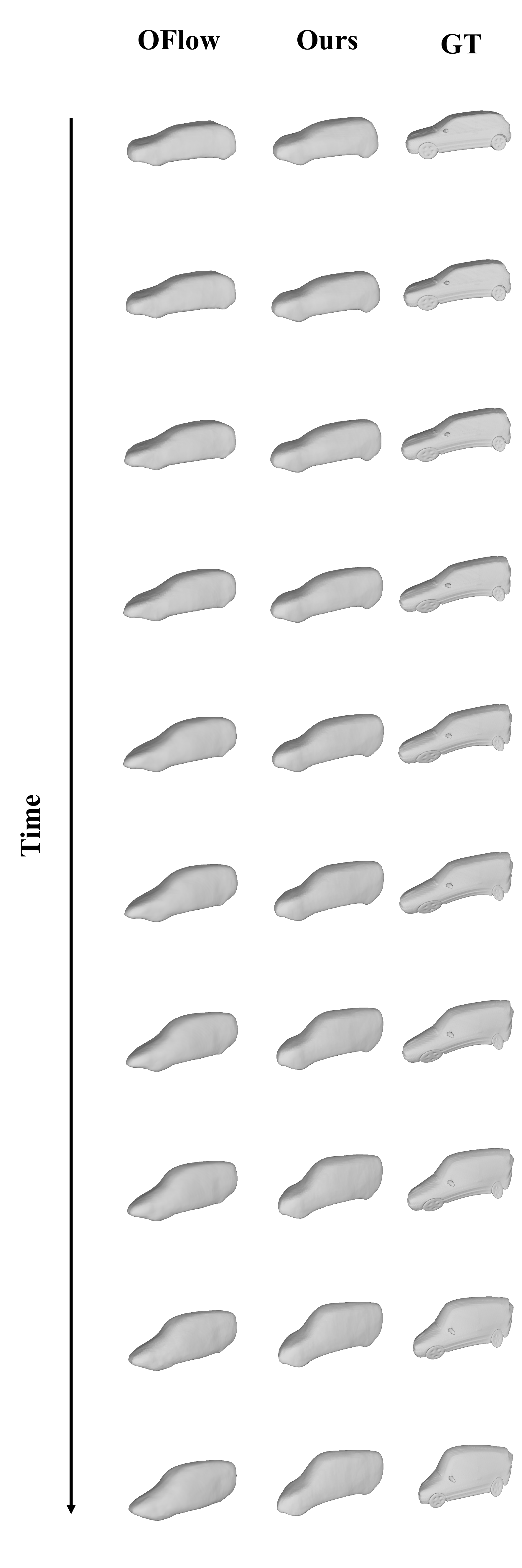}
\caption{\textbf{4D temporal completion (Warping Cars).}}
\label{fig:tempo_car} 
\end{figure}

\begin{figure}[tb]
\centering \includegraphics[width=1\linewidth]{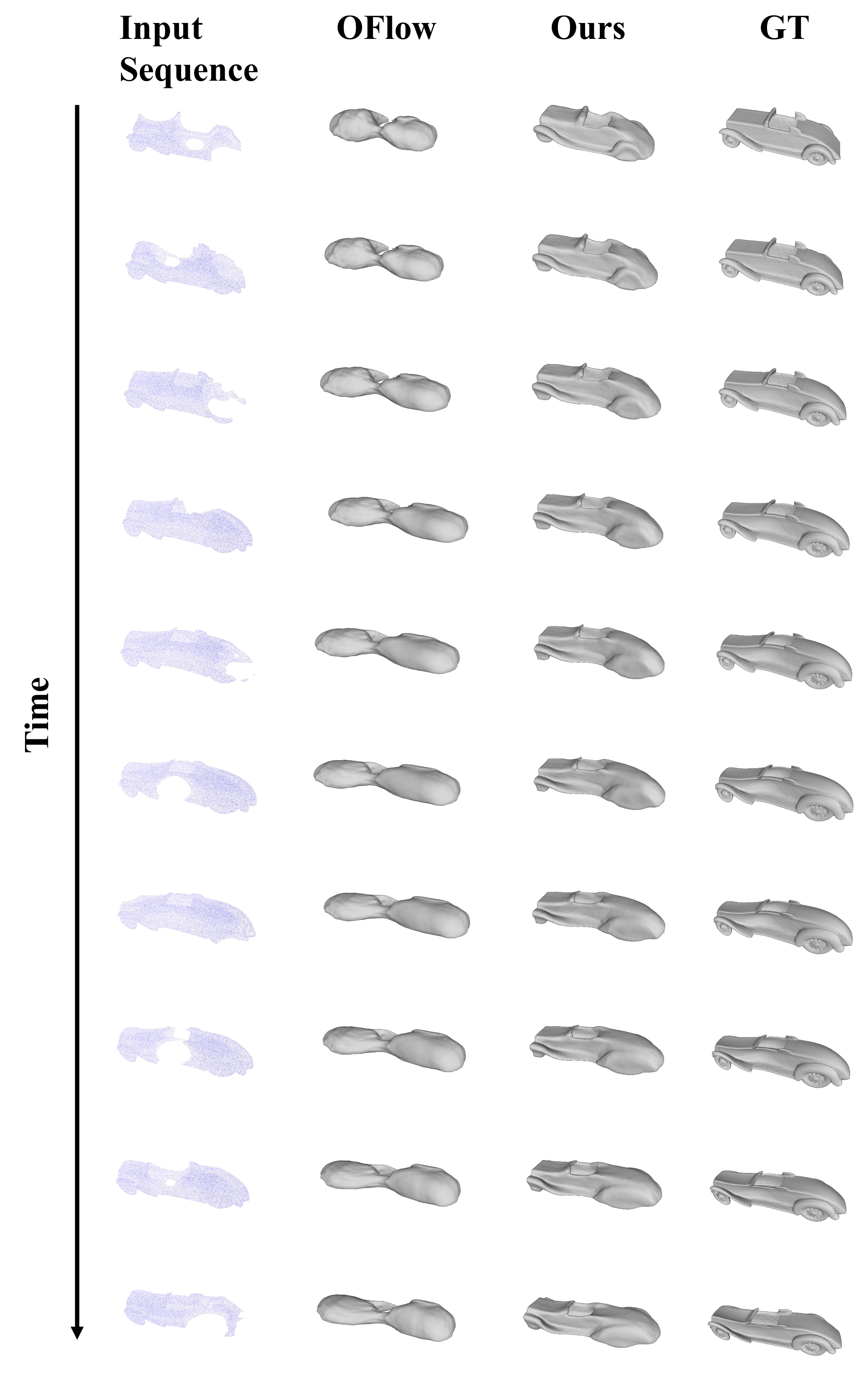}
\caption{\textbf{4D spatial completion (Warping Cars).} Note that we randomly remove points in the occupancy grid for optimization and we show the corresponding partial point clouds here for the convenience of visualization.}
\label{fig:spatial_car}
\end{figure}

\noindent \textbf{Pose and Motion Transfer} We evaluate the motion transfer performance of our method and OFlow on the Warping Cars dataset. Similar to that on the D-FAUST dataset, we choose 20 car shape and warping pairs and generate mesh sequences of length $L=17$ for evaluation.
The quantitative results are shown in Tab. \ref{tab:various_car}, and we shown a qualitative result in Fig. \ref{fig:transfer_car}. Our method obtains significantly better performance. OFlow gets unsatisfactory transfer results due to the inconsistency between the initial pose of the identity sequence and the motion sequence. Thanks to the compositional representation which disentangles pose from shape properly, our model successfully transfers the motion to a new car shape. The results on non-human dataset also verify the potential of our model for motion transfer task on objects from various categories.

\noindent \textbf{4D Completion} We also conduct the 4D completion experiment for Warping Cars.
Similar to the experiments on the D-FAUST dataset, we divide this task into two parts -- temporal completion and spatial completion. We select 18 point cloud sequences in the testing set, each of length $L=20$, and the strategies of removing frames and points are same as the previous experiments on the D-FAUST dataset, which are described in Section 4.4 of the main paper.

As the results shown in Tab. \ref{tab:various_car}, Fig. \ref{fig:tempo_car} and \ref{fig:spatial_car}, the proposed method achieves better results on both completion tasks than OFlow. We found that our method is more stable than OFlow during the completion experiments. The performance of OFlow heavily relies on the result of the first frame, because it only reconstruct mesh at $t=0$, and then use a Neural ODE to transform the positions of the points on the reconstructed mesh. When the result of the first frame is unsatisfactory, it is difficult for OFlow to have good shapes for subsequent frames, as shown in Fig. \ref{fig:spatial_car}. Our method applies the Neural ODE to update the latent pose code and reconstructs 3D model at each time step, which makes our results more stable.

\begin{figure}[htb]
\centering \includegraphics[width=0.91\linewidth]{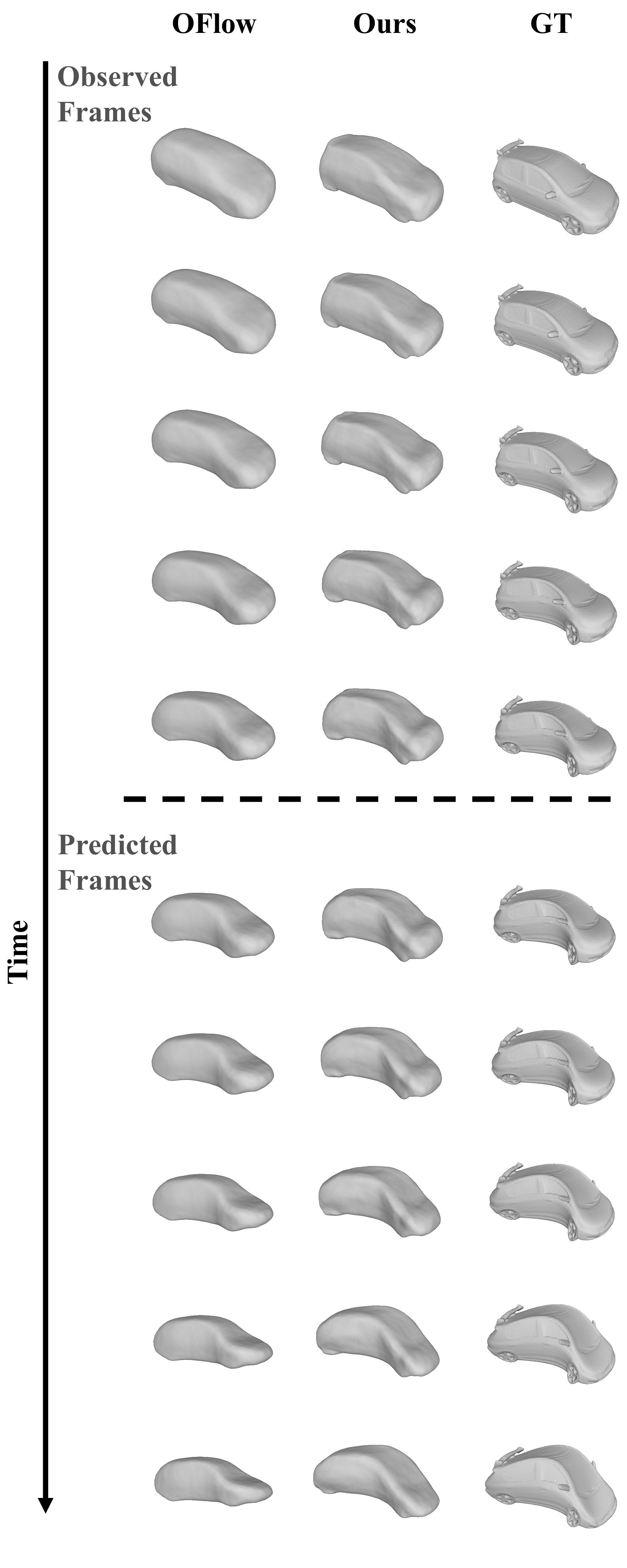}
\caption{\textbf{Future prediction (Warping Cars).} We remove the last 10 frames of the test sequence to investigate the extrapolation ability of our method. The results above the dotted line are reconstructions for partial observation, and the results below are future predictions.}
\label{fig:future_pred_car} 
\end{figure}

\noindent \textbf{Future Prediction}{\label{future_pred}}
In this experiment, we investigate the ability of our framework to predict future motion on our generated Warping Cars dataset. Same data for 4D completion task are taken, but we always remove the last 10 frames instead of randomly selected ones. We use the same hyper-parameters and optimization method based on back-propagation as the completion experiment. The quantitative
results are shown in Tab. \ref{tab:various_car} and the qualitative results can be found in Fig. \ref{fig:future_pred_car}. As shown, our method is capable of tracking existing observations and predicting more accurate future motion than OFlow.


\section{Motion Transfer with Different Initial Poses} 
\begin{figure}[tb]
\centering \includegraphics[width=0.95\linewidth]{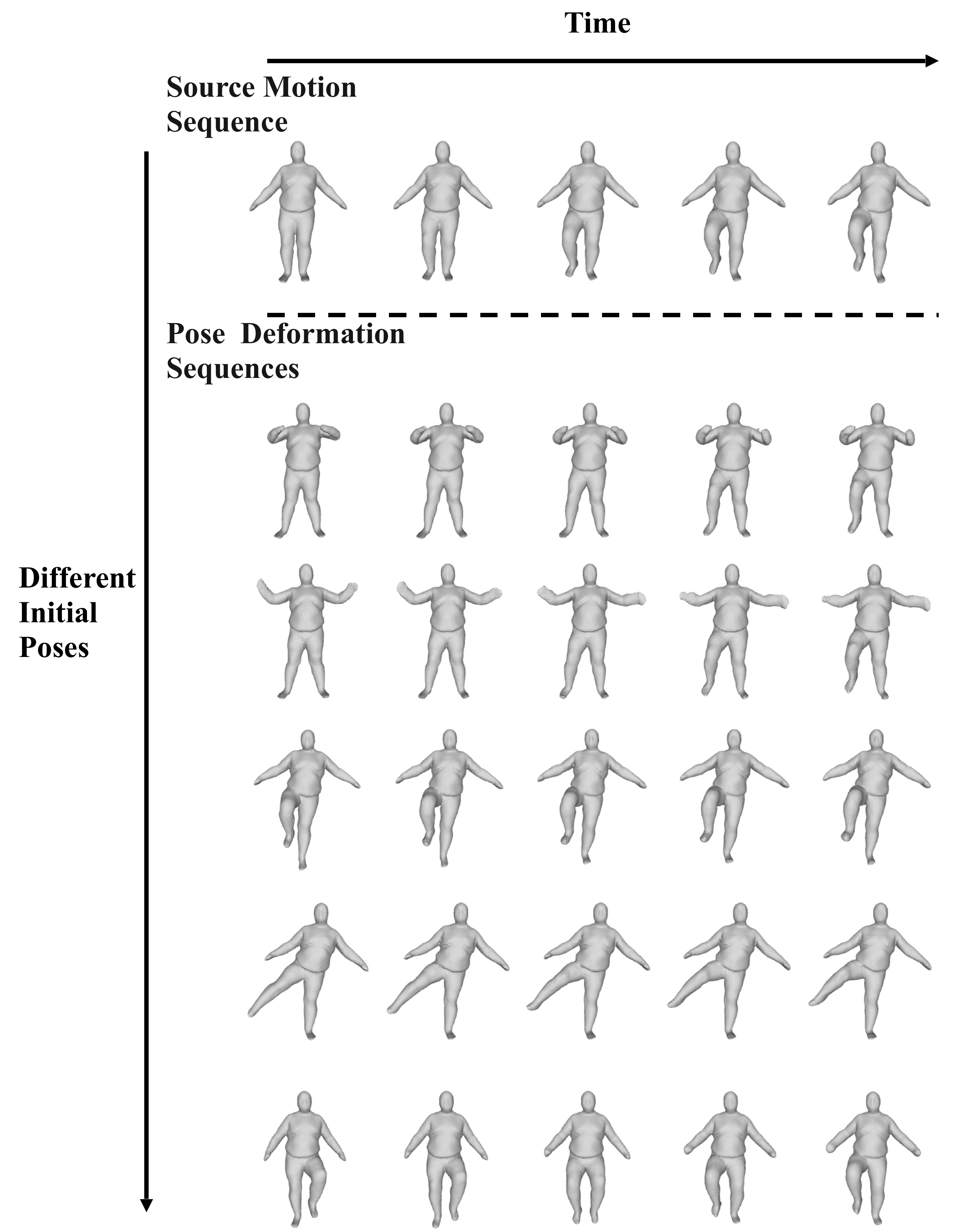}
\caption{\textbf{Motion transfer with different initial poses.}}
\label{fig:transfer_diff_pose} 
\end{figure}

In the previous motion transfer experiment, we transfer the motion code together with the initial pose code. To investigate if the motion code can be transferred without the initial pose code, we conduct an experiment that transfers a motion to different initial poses. The results are shown in Fig. \ref{fig:transfer_diff_pose}. Specifically, first, we use our motion encoder to obtain the source motion code from the motion sequence shown in the first line of Fig.~\ref{fig:transfer_diff_pose}. Then five mesh models with different poses are selected, and we use our identity encoder and pose encoder to get the identity code and initial pose code for each model, which then are fed into our decoder together with the source motion code.

The sequences shown in the second to sixth rows are results after transferring. Applying a motion to a new pose is challenging and sometimes ill-defined (e.g. forcing a stand-up motion to start with a standing pose). Surprisingly, our model still produces reasonable results if the new pose is not too different from the original one, which shows some robustness of motion transfer against the initial pose.

\section{More Qualitative Comparisons to OFlow}
\begin{figure}[tb]
\centering \includegraphics[width=1\linewidth]{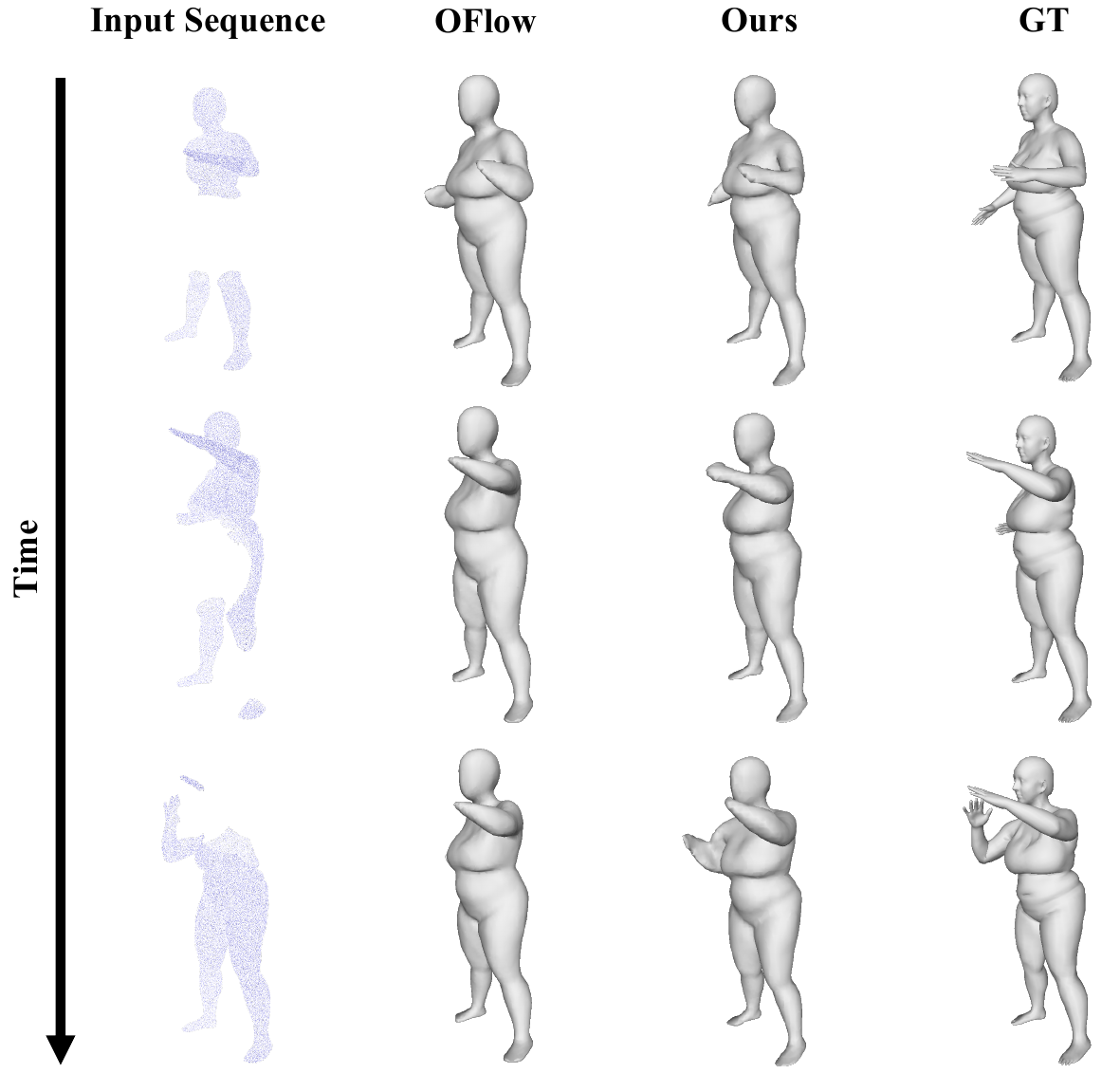}
\caption{\textbf{4D spatial completion (D-FAUST).} Note that we randomly remove points in the occupancy grid for optimization and we show the corresponding partial point clouds here for the convenience of visualization.}
\label{fig:spatial_dfaust}
\end{figure}

\begin{figure}[htb]
\centering \includegraphics[width=0.872\linewidth]{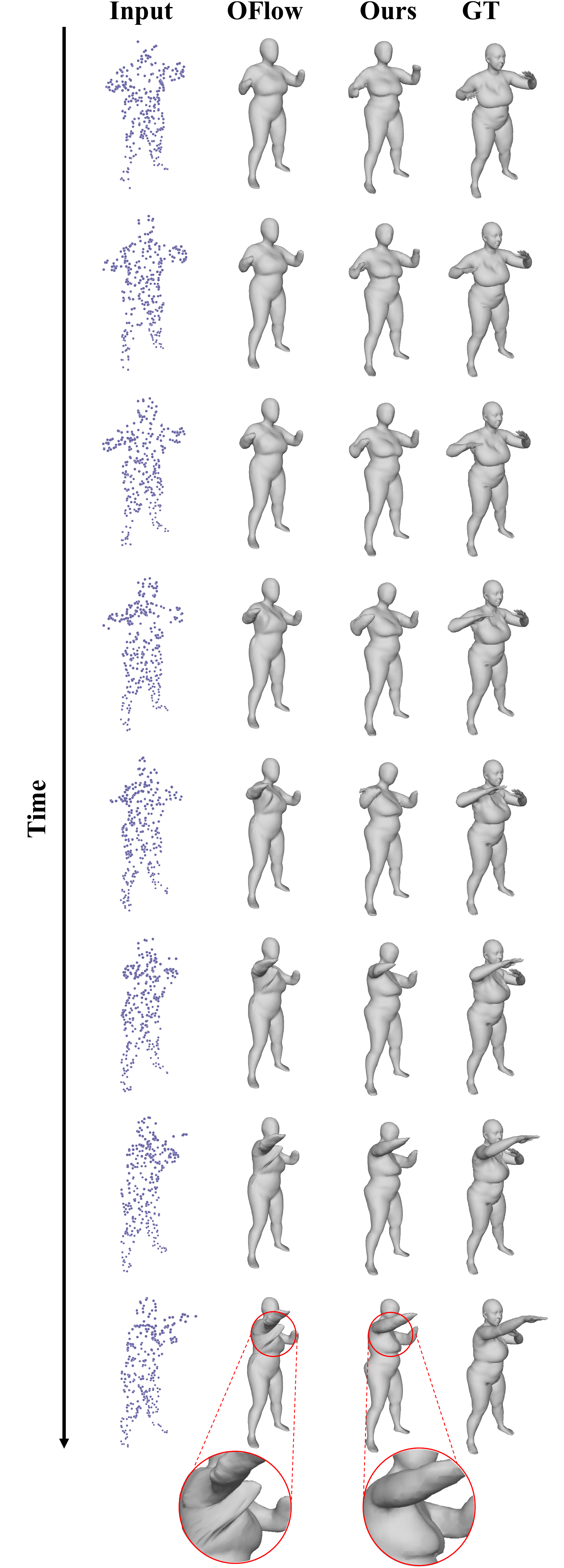}
\caption{\textbf{4D reconstruction from point cloud sequence (D-FAUST).} We show the input, ground truth and outputs of OFlow and our method for 8 equally spaced time steps between 0 and 1.}
\label{fig:more_recon_human} 
\end{figure}

\begin{figure}[htb]
\centering
\includegraphics[width=0.905\linewidth]{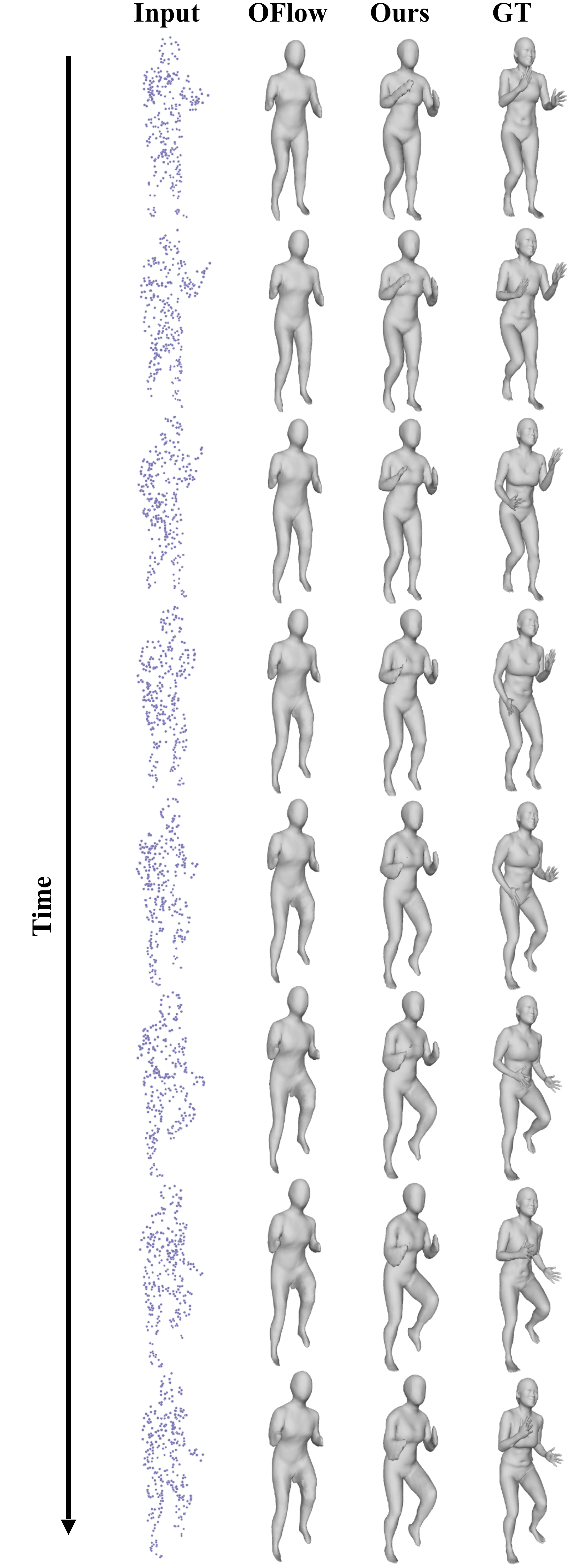}
\caption{\textbf{4D reconstruction from point cloud sequence (D-FAUST).} We show the input, ground truth and outputs of OFlow and our method for 8 equally spaced time steps between 0 and 1.}
\label{fig:more_recon_human2} 
\end{figure}

\begin{figure}[htb]
\centering \includegraphics[width=1\linewidth]{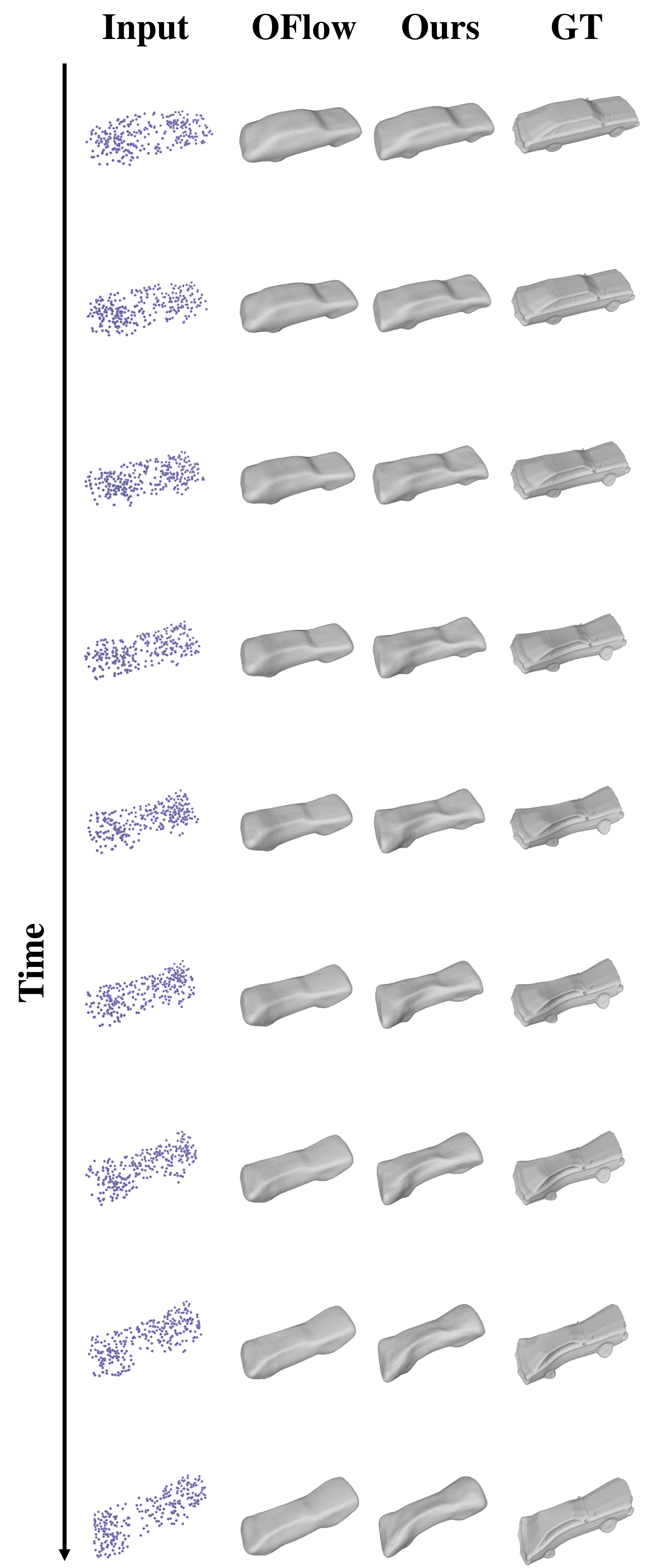}
\caption{\textbf{4D reconstruction from point cloud sequence (Warping Cars).}}
\label{fig:recon_car2} 
\end{figure}

\begin{figure}[htb]
\centering \includegraphics[width=1\linewidth]{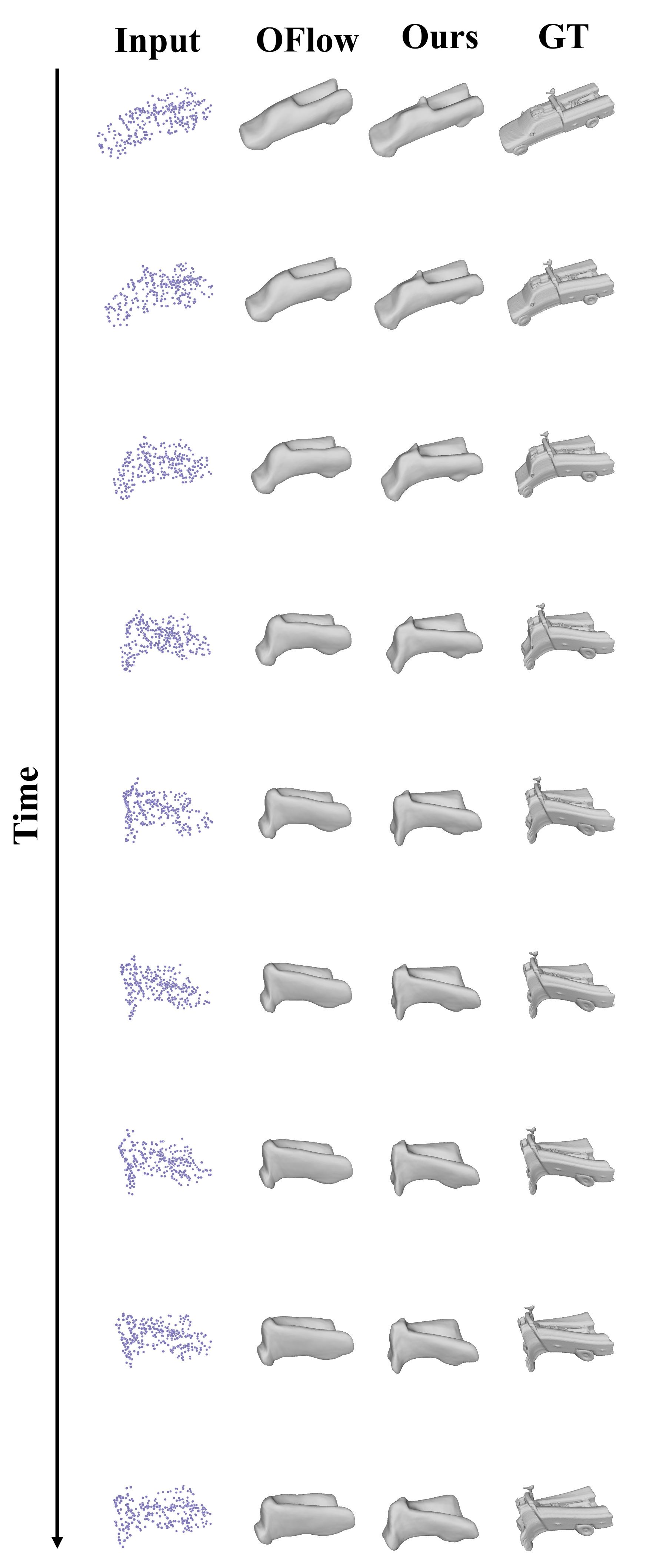}
\caption{\textbf{4D reconstruction from point cloud sequence (Warping Cars).}}
\label{fig:recon_car1}
\end{figure}

{\label{more_qailtative}}
We show more qualitative results of 4D spatial completion on the D-FAUST dataset in Fig. \ref{fig:spatial_dfaust}, and 4D reconstruction on both the D-FAUST and Warping Cars dataset in Fig. \ref{fig:more_recon_human}, \ref{fig:more_recon_human2}, \ref{fig:recon_car2} and \ref{fig:recon_car1}.


\end{document}